\documentclass[lettersize,journal]{IEEEtran}
\usepackage{amsmath,amsfonts}
\usepackage{algorithm}
\usepackage{array}
\usepackage{subcaption}
\usepackage{textcomp}
\usepackage{stfloats}
\usepackage{url}
\usepackage{verbatim}
\usepackage{graphicx}
\usepackage{booktabs}
\usepackage[table, dvipsnames]{xcolor}
\usepackage{tabularx}
\usepackage{multirow}
\usepackage{flushend}

\usepackage{cite}
\usepackage{soul}

\usepackage{siunitx}
\DeclareSIUnit{\nothing}{\relax}

\newcommand{\Nf}{\ensuremath{N_{\mathrm{f}}} }
\newcommand{\Nh}{\ensuremath{N_{\mathrm{h}}} }
\newcommand{\No}{\ensuremath{N_{\mathrm{o}}} }
\newcommand{\EffMetric}[1]{\ensuremath{\mathit{Eff}_{\mathrm{#1}}}}
\newcommand{\EffInfMetric}{\EffMetric{\text{inf}}}
\newcommand{\EffTrMetric}{\EffMetric{\text{tr}}}

\newcommand{\Zynq}{Zynq-7000 }%

\newcommand{\mylabfull}{Labo. de l'Int\'egration du Mat\'eriau au Syst\`eme}
\newcommand{\myorgfull}{Univ. Bordeaux, Bordeaux INP, CNRS}
\newcommand{\myaffilfull}{%
    \mylabfull, \myorgfull - F-33400 Talence, France
}

\begin{document}

\title{Conjoint Audio-to-Spikes Encoding and Processing \\ for Efficient Neuromorphic Speech Recognition}

\author{
    Valentin M. Meunier$^1$, Amélie Gruel$^{1}$, Pierre Lewden$^1$, Adrien F. Vincent$^1$ and Sylvain Saïghi$^{1,*}$ \\
    \small{$^1$\myaffilfull} \\
    \small{$^*$Author to whom any correspondence should be addressed. --- \texttt{\{firstname.surname\}@u-bordeaux.fr}}
}

\markboth{TNNLS}{TNNLS}

\maketitle

\begin{abstract}
    Obtaining data from neuromorphic sensors and processing it with Spiking Neural Networks is a promising solution to lower the energy cost of artificial intelligence.
    The current rarity of natively neuromorphic datasets promotes the development of software tools to translate input sensory data into spikes.
    However, highly bio-mimetic simulators can be challenging to implement on digital hardware.
    In this work, we evaluate the neuromorphic encoding and subsequent classification of audio into spikes using a non-learnable, high-level, programmable encoder targeting hardware implementation on FPGA. We quantify the pipeline's efficiency with hardware-agnostic metrics based on the quantitative spiking activity. Our study focuses on the simultaneous optimisation of encoder and classifier: the first provides efficient and informative data so that the latter achieves a better performance with an overall lower energy cost at learning and inference. 
    This work introduces the first end-to-end neuromorphic spike-encoding and evaluation of the TIMIT dataset. Our simple feedforward network reaches a
    classification accuracy of 99.77\% on a spike-encoded Heidelberg Digits, overcoming the neuromorphic state of the art on this benchmark dataset.
\end{abstract}


\section{Introduction}
\label{sec:introduction}

Neuromorphic computing emerged in the 1980s, combining electrical engineering, computer science, mathematics and neuroscience to efficiently implement a bio-inspired artificial intelligence~\cite{Mead2020}. Since Mead's early work on silicon retina and cochlea~\cite{Mead1988}, this field of research has been thriving, reinforced by the work of many scientists leading to the creation of diverse sensors and processing strategies.
Neuromorphic sensors record and process sensory data as spikes sequences~\cite{Vanarse2016,Yang2023}, leading to the creation of several event-based datasets such as N-MNIST~\cite{N-MNIST}, DVS-Gesture~\cite{DVSGesture}, GEN1~\cite{Gen1}, etc. These datasets remain however rare compared to conventional datasets, especially in other modalities than vision. To mitigate this scarcity, software simulators have been developed to emulate neuromorphic sensors and synthesise event streams from standard inputs, especially video~\cite{v2e,vid2e,v2ce}.

Neuromorphic audio-to-spikes encoding has seen a boost of interest since the introduction in 2020 of the bio-mimetic encoder Lauscher~\cite{SHD}, an artificial cochlea model reproducing aspects of the inner ear and ascending auditory pathway. The authors recorded and published in~\cite{SHD} the novel, conventional audio dataset Heidelberg Digits (HD) and its neuromorphic alternative, obtained using the conversion chain Lauscher and renamed ``Spiking Heidelberg Digits'' (SHD). From the conventional audio dataset Google Speech Commands v0.2 (GSC)~\cite{Warden2018Speech}, they similarly encoded and made available a spiking version designated ``Spiking Speech Commands'' (SSC). Both SHD and SSC are now widely used  benchmarks for neuromorphic speech recognition. However, although highly bio-mimetic, Lauscher is computationally demanding and energy-intensive; tuning its parameters for task-specific needs can be challenging due to the intricate parametrisation impacts on spikes statistics. Furthermore, its mathematical complexity and memory footprint hinder efficient digital hardware implementation, especially on embedded targets. 

Alternatives emerged with a lighter computational cost -- such as Speech2Spikes~\cite{speech2spikes} or Spiking-Leaf~\cite{Song2024Spiking} -- or directly targeting hardware~\cite{EncodingSurvey2021, SurveyAudio2025, lee2002speaker}, but few encoders effectively bridge bio-inspiration, computational efficiency, and hardware feasibility. One of them is the High-Level Programmable strategy (referred to as ``HLP'' in the remainder of this work) we introduced in~\cite{meunier_aicas2025}: this neuromorphic filterbank encoder was extended from~\cite{Papierclemenceadrien}, a work specifically targeting implementation of audio-to-spikes conversion on Field-Programmable Gate Arrays (FPGAs). The programmable logic and routing resources of the latter enable favourable trade-offs in area, speed, and power. The HLP encoder effectively balances out bio-inspiration and ease of implementation -- thus anticipating  the longer-term emergence of natively event-based, low-power bio-inspired audio sensors such as~\cite{Chan2007AEREAR}, while delivering short-term, practical efficiency benefits on current platforms. \\

We thus set our sight on this HLP strategy to investigate the impact of audio-to-spikes encoding on the classification performance and energy efficiency of a downstream Spiking Neural Network (SNN). More bio-mimetic than conventional artificial neural networks, SNNs mimic the dynamics of the biological neuron and similarly operate on sparse, event-driven signals~\cite{maass1997networks}.
Their low energy use can be estimated using hardware-agnostic efficiency metrics, which measure the trade-off between performance (here, the classification accuracy) and the number of synaptic operations required over the whole SNN, approximating its theoretical dynamic energy usage~\cite{sorbaro2019}.
This paper extends previous surveys emphasising the critical impact of encoder design in the neuromorphic audio domain~\cite{EncodingSurvey2021,SurveyAudio2025}, and further argues the necessity for conjoint optimisation of spike generation and processing. 

The remainder of this work successively presents the neuromorphic pipeline and its parametrisations used in our comparative study, detailing the adopted HLP strategy and associated SNN classifier and highlighting their differences with other approaches in the literature; the metrics and datasets used to validate this pipeline; and the various optimisation steps pursued before the final evaluation on software.
Finally, we discuss the implementation of this neuromorphic pipeline on FPGA, from LLP encoding (for Low-Level Programmable -- a direct adaptation to hardware of our HLP strategy) to SNN classification. We analyse in this work the first audio-to-spikes encoding and conjoint SNN evaluation of the traditional dataset TIMIT~\cite{TIMIT} -- indeed previous works have studied the SNN classification of a non-spiking TIMIT~\cite{dong2018unsupervised}; or the encoding of TIMIT into spikes without any subsequent processing by an SNN~\cite{chakrabartty2010exploiting, li2012real, pan2020efficient}. 
Overall, our best simulated results reach a 99.77\% accuracy on the spiked HD dataset thus outperforming the state of the art and FPGA implementation of a more restricted architecture reaches a \qty{99.63}{\percent} accuracy on the English subset of HD.

\begin{figure*}[t]
    \centering
    \includegraphics[width=\textwidth]{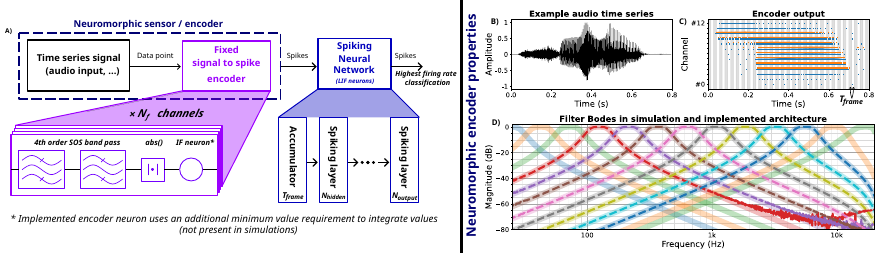}
    \caption{\textbf{A)} Our general architecture for SNN classification of audio time-series. 
    Our neuromorphic programmable encoder converts time-series into spikes using a bank of $\Nf$ band-pass Butterworth filters in their Second Order Section (SOS) form, which output is rectified then fed to a spiking neuron.
    The Accumulator block accumulates encoded spikes into a specific number of computation steps then fed to our SNN classifier, thus reducing the number of computations performed by the latter. 
    The classification identifies the winner as the output neuron producing the highest spike count.
    \textbf{B)} Example of an audio time-series of label ``zero'' from the HD dataset.
    \textbf{C)} Raster plot of the spiking activity of each channel of the encoder, where the input is the audio sample shown in B. Each vertical line is a spike: blue spikes were obtained in simulation out of $\Nf=13$ encoding channels; orange spikes were obtained on hardware, through the architecture implemented on FPGA, out of $\Nf=8$ central channels (see further details on hardware implementation in Sec.~\ref{sec:hardware}). Simulation and hardware encoder have qualitatively similar outputs.
    \textbf{D)} Logarithmic distribution of filters used to produce the spikes in C, with a sampling rate of \qty{48}{\kilo\hertz}. Thick lines correspond to our HLP encoder, i.e. filters in the human hearing range; dotted lines to our LLP encoder.}
    \label{fig:encoder_wide}
\end{figure*}

\section{Methods}

\subsection{Scope of the analysis}
\label{subsec:scope}

This work studies an efficient neuromorphic pipeline for audio classification and intended for FPGA implementation, comprising a bio-inspired, non-learnable cochlear encoder and an SNN classifier. 
This approach is benchmarked solely against similar, end-to-end neuromorphic pipelines, as detailed in the paragraphs below. 
We focus our analysis on optimising the audio-to-spikes encoding to produce both informative and efficient data for downstream SNN training and inference.\\

\paragraph{Regarding learning}
Both encoder and classifier contain spiking neurons (see Sections~\ref{subsec:encoder} and \ref{subsec:snn}), but only the ones within the classifier are trained to process the given spikes. No learning takes place within the conversion of audio waveforms into spikes: this ensures the encoder reusability across tasks and datasets, while allowing the downstream insertion of interchangeable classifiers for task-specific learning. 

We will thus remove from our analysis pipelines involving a learnable front-end, such as Spiking-LEAF~\cite{Song2024Spiking}.
This work optimises the data encoding conjointly with the back-end SNN classifier, leading to a 93.95\% accuracy on GSC. The benefits of its high performance are downplayed by the non generalisability of the encoding, tailored to a particular task and dataset.
\\

\paragraph{Regarding non-spiked input data}
As publicly available spiking datasets are scarce, many approaches feed non-spiking data directly to downstream SNN classifiers. In this case, the first layer usually receives floats as input, then produces spikes in response. Another solution converts non-spiking data into spikes by interpreting each normalised feature of the signal as a per-timestep probability of firing; spikes are then sampled independently over a specified number of steps, thus obtaining a final spike count proportional to the normalised feature (see the spike generator implemented in~\cite{eshraghian2021training}). This probabilistic process adds latency and energy cost to the direct emission of event-streams performed advantageously by a neuromorphic cochlear encoder.

This work does not consider as benchmark pipelines taking non-spiked data as input. Indeed, there is a qualitative difference between learning from spiked and non-spiked inputs (e.g., from SSC and GSC). In the latter case, the encoding is often optimised jointly with the downstream SNN for a specific task, thus yielding higher accuracy than processing a fixed spike encoding. For example, SNN classifiers trained on GSC (i.e. on raw audio) report up to 96.23\% accuracy at inference~\cite{bal2025pspikessm}, whereas the best reported SNN classification accuracy is 83.69\% on SSC~\cite{wang2024spikescr}, i.e. on spikes fixed by the Lauscher encoder.

Furthermore, classifying audio waveforms \textit{versus} spiked data is ultimately not the same task. Pipelines carrying out SNN classification on non-spiking inputs primarily seek bio-inspired efficiency only at the processing stage, or address the lack of spiking datasets. In contrast, the pipelines we analyse in this work are end-to-end neuromorphic: the data is efficiently encoded and stored as spike trains and can be reused by multiple downstream SNNs, as the encoding is not specialised for a particular network or task.

Results obtained in~\cite{speech2spikes} fall within the scope of our study as it makes use of the Speech2Spikes encoder to produce spikes from audio, and classifies them using a SNN. However, it is to be noted that Speech2Spikes is not neuromorphic \textit{per se}, but relies on non-neuromorphic operations such as Fourier transforms to produce spiked data. \\

\paragraph{Regarding non-neuromorphic classifiers}
Finally, the scope of our study excludes non-neuromorphic classifiers (i.e. conventional artificial neural networks), as they were demonstrated times again as less efficient than SNNs~\cite{eshraghian2021training}. Additionally, comparing SNN with conventional approaches is unfair to the inherent continuity of the latter. This allows better optimisation than for SNNs, as demonstrated in~\cite{schone2024scalable} which achieves the state-of-the-art performance of 88.4\% on GSC using a modern, non-spiking recurrent deep state-space model. \\

\paragraph{Regarding the task}
We target a general-purpose speech pipeline, rather than one narrowly tailored to a single dataset. We keep the preprocessing and input representation identical across datasets, aiming to reduce the hyperparameters of the pipeline in need of tuning, and avoid dataset-specific heuristics (e.g., keyword-dependent augmentations or speaker-ID tricks). This design aims to prevent the pipeline over-specialisation for specific datasets, in order to generalise our HLP encoder tuning on different human-speech datasets.

\subsection{HLP cochlear encoder}
\label{subsec:encoder}

\paragraph{Overall mechanism}

The HLP cochlear encoder used in our pipeline targets efficient hardware realisation while retaining bio-inspiration for efficiency. It is implemented according to the following common structure: each canal comprises a band-pass filter, a rectifier and a simple spiking neuron. The model adopted for the latter is the Leaky Integrate-and-Fire (LIF) one~\cite{LIF}, as many recent works have demonstrated its ability and relevance in cochlear encoding, for task-specific information retention~\cite{ElFerdaoussi2022,EffSpikeSpeech2022,HTSpike2025}.
The rectification step consists in feeding to each LIF neuron the absolute value of its upstream filter output.
An additional preprocessing step can be applied to raw audio signal, by scaling its amplitude by the root-mean-square (RMS) level to decrease the sensitivity to outliers in noisy datasets, similarly to~\cite{cramer2019heidelberg}.

The HLP encoding is driven using a small set of parameters that directly shape output spike density, facilitating its tuning. This simple design improves feasibility on embedded FPGAs, especially compared to Lauscher’s higher-order, memory-intensive model; the latter encodes data by resolving complex equations of the physical propagation of the sound along the cochlea membrane.\\ 

\paragraph{Parametrisation}
In this work, audio is encoded into spikes using \Nf Butterworth filters, logarithmically-spaced~\cite{audioOverview} within the human hearing range~\cite{hearingRange}.
More specifically, the central frequency of the lowest filter is common to all considered datasets, and set to \qty{25}{\hertz}. However, as the sampling rate differs from one dataset to another, the central frequency of the highest filter $F_{\max}$ is set to \qty{16667}{\hertz} for HD and, differently, to \qty{6650}{\hertz} for GSC and TIMIT.
The choice of these $F_{\max}$ values places the cut-off frequency of the highest filters just below the Nyquist frequency, i.e. half the value of the sampling rate of the audio. 
Fig.~\ref{fig:encoder_wide} illustrates such a distribution of filters, where $\Nf = 13$. 

In addition, each band-pass filter is characterised by its quality factor $Q$ and its order. The quality factor $Q$ controls the filter's ability to select frequencies, i.e. its bandwidth: larger $Q$ yields a narrower pass-band and a longer, more oscillatory impulse response. The filter order corresponds to the number of poles in a filter~\cite{filter_order}. It thus impacts the number of time steps involved in the computation and the associated memory footprint, where a higher order results both in a steeper roll-off and an increase in the number of recursive states. 

In addition to the filter parameters, the pipeline heavily depends on the setting of two key parameters that shape the output spike density: the firing threshold and the membrane decay rate of the LIF neurons involved in the encoding. 
Each set of encoding parameters' values corresponds to a new neuromorphic variant of the original non-spiking dataset. \\

\subsection{SNN classifier}
\label{subsec:snn} 

The audio classification part of our neuromorphic pipeline is performed by a feedforward, fully-connected SNN, implemented in software through the snnTorch library~\cite{eshraghian2021training} if not stated otherwise. Each layer performs an affine linear mapping learning both weights and additive biases; followed by a LIF neuron, which decay rate $\beta$ is set to 0.9 and which membrane potential is reset by subtraction.
Network outputs were obtained by counting spikes on the output units across time and selecting the class with the largest spike count.

Training uses an arctangent surrogate gradient function to enable backpropagation through time relying on the Adam optimiser~\cite{adamoptimiser}. The latter is combined with a Mean Square Error Spike Count Loss to perform a rate-based decoding of the spiking activity, where the target spike rates are corrected to limit the dead neuron problem~\cite{Shrestha2018}. To ensure robustness and statistical reliability, each run was repeated at least three times using distinct random seeds. Initial and final network weights and biases were saved systematically, providing a thorough basis for subsequent hardware implementation. The accuracy reported in the remainder corresponds to the best inference accuracy obtained over the specified number of epochs, after averaging over the multiple runs.

The classifier's architecture consists of one input layer of \Nf units; one to three hidden layers of same number \Nh neurons; and one output layer of \No neurons. \Nf corresponds to the number of filters used by the cochlear encoder: Lauscher uses $\Nf = 700$~\cite{SHD} and Speech2Spikes uses $\Nf = 80$~\cite{speech2spikes}; in this work, we vary \Nf in the HLP encoder to study its impact on downstream classification performance.
$\No$ corresponds to the number of classes within the targeted dataset -- either 20, 35 or 39 (see Table~\ref{tab:datasets}).

We record the number of synaptic events produced within the model at each layer, both during training and inference. This spike count highly varies depending on the SNN classifier and the dataset, but also on the encoding and its parameters.

\subsection{Datasets}
\label{sec:datasets}

\begin{table*}[ht]
    \centering
    \caption{Statistics of the three datasets under study.
    }
    \label{tab:datasets}
    \begin{tabular}{l||c|c c c c c}
        \hline
        \textbf{Dataset} & \textbf{Encoder} & \textbf{Sampling rate} & \textbf{\# labels $\No$} & \textbf{\# speakers} & \textbf{\# samples} & \textbf{Split} (train~/~valid~/~test) \\
        \hline
        \multirow{2}{*}{HD} & Lauscher (SHD) & \multirow{2}{*}{\qty{48}{\kilo\hertz}} & \multirow{2}{*}{20} & \multirow{2}{*}{12} & \multirow{2}{*}{\qty{10420}{}} & \qty{78}{\percent}~/~-~/~\qty{22}{\percent} \\
         & HLP (NHD) & & & & & \qty{80}{\percent}~/~-~/~\qty{20}{\percent} \\
        \hline
        \multirow{2}{*}{GSC} & Lauscher (SSC) & \multirow{2}{*}{\qty{16}{\kilo\hertz}} & \multirow{2}{*}{35} & \multirow{2}{*}{\qty{2618}{}} & \multirow{2}{*}{\qty{105830}{}} & \qty{71}{\percent}~/~\qty{10}{\percent}~/~\qty{19}{\percent} \\
         & HLP (NSC) & & & & & \qty{72}{\percent}~/~\qty{9}{\percent}~/~\qty{19}{\percent} \\
        \hline
        TIMIT & HLP (NTIMIT) & \qty{16}{\kilo\hertz} & 39 & \qty{630}{} & \qty{241225}{} & \qty{80}{\percent}~/~-~/~\qty{20}{\percent} \\
        \hline
    \end{tabular}
\end{table*}

We apply our end-to-end neuromorphic pipeline for audio classification to three benchmark datasets. To prevent confusion with other spiked versions of those three, raw audio datasets, we add the qualifying adjective ``Neuromorphic'' to the initial dataset named, shortened into ``N*'' (similarly to the ``Spiking'', shortened into ``S*'' used in~\cite{SHD}).
We compare several variants of the encoded datasets by sweeping the HLP encoder's parameters (see Sec.~\ref{subsec:encoder}), and we provide the produced spikes accumulated over 100 computation steps as input to the SNN classifier.
For example, the accumulation into 100 computation steps of HD's \qty{1.4}{\second}-long samples at a \qty{48}{\kilo\hertz} sampling frequency, produces frames of $T_{frame}= 1.4 \times 48000 / 100 = 672$ sampling points. \\

\paragraph{Heidelberg Digits (HD)} 
This raw audio dataset, introduced in~\cite{SHD}, consists of high-quality aligned studio recordings of digits from 0 to 9 spoken in both German and English. Twelve distinct speakers were recorded, two of whom are only present in the test set.
Very few samples last longer than \qty{1.4}{\second}. We standardise all samples to \qty{1.4}{\second} (through either clipping or zero-padding) similarly to~\cite{SHD}.
The sampling rate is \qty{48000}{\hertz}. 

We compare our spiked encoding of HD (i.e. NHD) and its performance to other state-of-the-art approaches falling within the scope of our work (see Sec.~\ref{subsec:scope}), all listed in Tab.~\ref{tab:sotapipeline}. These approaches use either the Lauscher encoder (i.e. the SHD dataset)~\cite{SHD}, or the Speech2Spikes one (i.e. the S2S-HD)~\cite{speech2spikes}, on which are applied various SNN classifiers.\\

\paragraph{Google Speech Commands v0.2 (GSC)} Introduced in~\cite{Warden2018Speech}, this large dataset consists of one-second long utterances of 35 short words (including the digits 0 to 9 spoken in English). In total, 2,618 distinct speakers contributed to the dataset by recording their samples on a publicly available website.
The quality of the audio thus highly varies from one sample to another.
The ``noisiness'' variations between samples (wrt. background noise, dynamics, etc.) is reduced by applying RMS scaling on the recorded signals (see Sec.~\ref{subsec:encoder}).
The sampling rate is \qty{16000}{\hertz} for all samples.

Similarly to HD (see above), we compare NSC and its performance to approaches falling in our scope, i.e. applying SNN classification to SSC~\cite{SHD} or S2S-GSC~\cite{speech2spikes}. \\

\paragraph{TIMIT} This phoneme dataset introduced in 1993~\cite{TIMIT} contains sentences of a few seconds, composed of 61 labelled phonemes and spoken by 630 distinct speakers. It was recorded at a sampling rate of \qty{16000}{\hertz}.
We segment the original samples, initially corresponding to a sequence of multiple labels, into mono-label samples of \qty{0.25}{\second} each containing a single phoneme. Furthermore, we focus our study on a subset of the dataset, reduced to 39 phonemes out of 61 as done in~\cite{lee2002speaker}, in order to come closer to the classification task of GSC for easier comparison. We also apply RMS scaling to TIMIT signals, similarly to GSC.

Previous works have studied the SNN classification of a non-spiking TIMIT~\cite{dong2018unsupervised}; or the encoding of TIMIT into spikes without any subsequent process by an SNN~\cite{chakrabartty2010exploiting, li2012real, pan2020efficient}. To the best of our knowledge, this paper thus introduces the first  end-to-end neuromorphic encoding and processing of the TIMIT dataset. No published baseline exists to which we can compare our HLP approach; nonetheless, our evaluation of TIMIT remains quite informative. Firstly, it tests the reproducibility of our architecture tuning under a dataset that matches GSC in sampling rate (\qty{16}{\kilo\hertz}) and has a comparable number of classes (39 phonemes vs. 35 words). Secondly, it probes performance on shorter acoustic units -- phonemes rather than words -- thereby assessing our pipeline under reduced temporal context.
Finally, NTIMIT constitutes a novel resource for the neuromorphic community.\\

\subsection{Evaluation metrics} 
\label{subsec:metrics}  

The goal of this work is to simultaneously evaluate the classification performance, using its accuracy but also its energy cost, which heavily depends on the transfer of data~\cite{ENERGYDampfhoffer,NbrOperationDampfhoffer}.
There is currently no clear consensus in the neuromorphic community on the strategy to be used to report the energy cost of SNN models across heterogeneous neuromorphic platforms.
We use two hardware-neutral metrics of efficiency, that we define as the ratio between accuracy and two different spike counts.
Those efficiency metrics can be retrofitted to any target hardware by applying any user-chosen per-event costs (synaptic events, neuron updates, memory accesses, etc.).
In addition to making our results comparable and reproducible across datasets and devices, such efficiency metrics allow to more easily understand the impacts of HLP encoder's tuning on the energy cost of the neuromorphic pipeline.

The training efficiency $\EffTrMetric$ evaluates the energy spent on training in order to achieve a given accuracy. It is computed as the ratio between the test accuracy achieved at a given epoch $e$, divided by the sum of all synaptic operations produced within the classifier during its training phase, from the first epoch to the epoch $e$ (at which the accuracy is considered).

The inference efficiency $\EffInfMetric$ focuses of the energy cost during inference.
It corresponds to the test accuracy achieved at a given epoch $e$, divided by the number of spikes emitted within the SNN classifier during its inference phase (only) at epoch $e$ (at which the accuracy is considered). 

\section{Results}
\label{sec:Results} 

\subsection{Encoder optimisation} 
The following section studies the impact of HLP encoder optimisation on downstream classification. This analysis has been initiated in a previous study~\cite{meunier_aicas2025}, tuning the quality factor $Q$ of the filters as well as the membrane decay $D$ and threshold $T$ of the LIF neurons involved in each channel according to the classification results of a recurrent SNN model strongly inspired by~\cite{SHD}. The focus of this previous work was solely on optimising the encoding; the current one aims to conjointly focus tuning efforts on encoding and classification. \\

\subsubsection{Channel parametrisation}

In~\cite{meunier_aicas2025}, we carried out a grid search on the three hyperparameters $Q$, $D$ and $T$ on 75 variants of both the HD and GSC datasets, evaluating them using a recurrent SNN classifier. This study indicated that a low quality factor $Q$ was sufficient for good performance; similarly, a smaller decay $D$ led to a better performance. The authors also studied how far the threshold could be raised without hurting accuracy; higher threshold reduces emitted spikes in the datasets, which lowers the energy cost.

We have since validated these conclusions with our simpler and more hardware friendly SNN described in section~\ref{subsec:snn}, which we use for the remainder of this study. We observed that removing membrane decay in the encoder yielded the best downstream accuracy; and that, without decay, the tuning of the threshold was enough to control spike density. Removing the decay of the neurons of our encoder simplifies the tuning of our pipeline and its implementation on hardware, as our HLP encoder now requires Integrate-and-Fire (IF) neurons~\cite{gerstner2002snm} instead of LIF neurons. For HD and GSC, setting the threshold to 1 achieved higher accuracy with fewer spikes than lower threshold values; for TIMIT, the best trade-off occurred for a threshold $T=0.06$. These values of hyperparameters are used in all subsequent experiments.

\begin{figure}[!htbp]
    \centering
    \includegraphics[width=\linewidth]{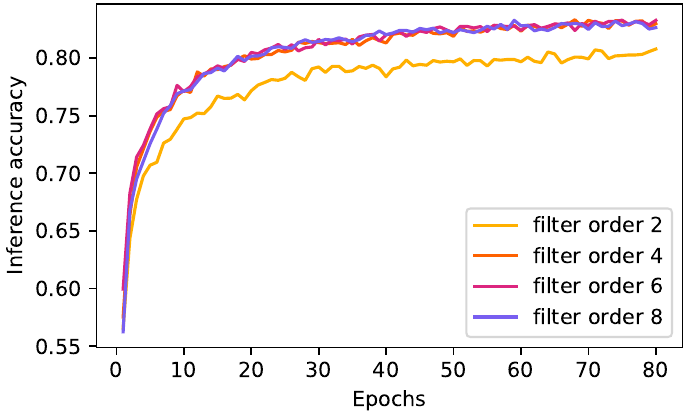}
    \caption{Impact of the encoding filter order on downstream classification performance. The inference accuracy is averaged over 3 runs, using an architecture 8-100-100-10 applied to the NSC-digit subset (i.e. only samples corresponding to digits within NSC). All encoding parameters other than the filter order are set according to Tab.~\ref{tab:HLP_encoder_param}.}
    \label{fig:ComparisonencodingFilterORDER}
\end{figure}

Another channel parameter tuned in this work is the order of the filter, knowing that the cost of its implementation is proportional to its complexity. Fig.~\ref{fig:ComparisonencodingFilterORDER} shows that there is no significant impact on downstream classification of NHD using filters of the fourth order or above -- while we do observe a lower accuracy for an order 2. We thus parameterise all filters with an order 4 in the remainder of this work. All channel parameters are summed up in Tab.~\ref{tab:HLP_encoder_param}. \\

\begin{table}[h]
    \centering
    \caption{
        HLP encoder parameters for best accuracy.
        }
    \begin{tabular}{c|c c c}
        \hline
        Parameter & NHD & NSC & NTIMIT  \\
        \hline
        Quality factor $Q$ & 2.5 & 2.5 & 2.5 \\
        Membrane decay $D$ & 0 & 0 & 0 \\
        Threshold $T$ & 1 & 1 & 0.06 \\
        Order & 4 & 4 & 4 \\
        RMS scaling & / & 0.02 & 0.02 \\
        $F_{\min}$ & \qty{25}{\hertz} & \qty{25}{\hertz} & \qty{25}{\hertz} \\
        $F_{\max}$ & \qty{16667}{\hertz} & \qty{6650}{\hertz} & \qty{6650}{\hertz} \\
        \hline
    \end{tabular}
    \label{tab:HLP_encoder_param}
\end{table}

\begin{figure*}[!t]
    \centering
    \begin{subfigure}[c]{\textwidth}
        \caption{Comparison between NHD variants and SHD.}
        \label{fig:ComparisonFilterNHD}
        \includegraphics[width=\textwidth]{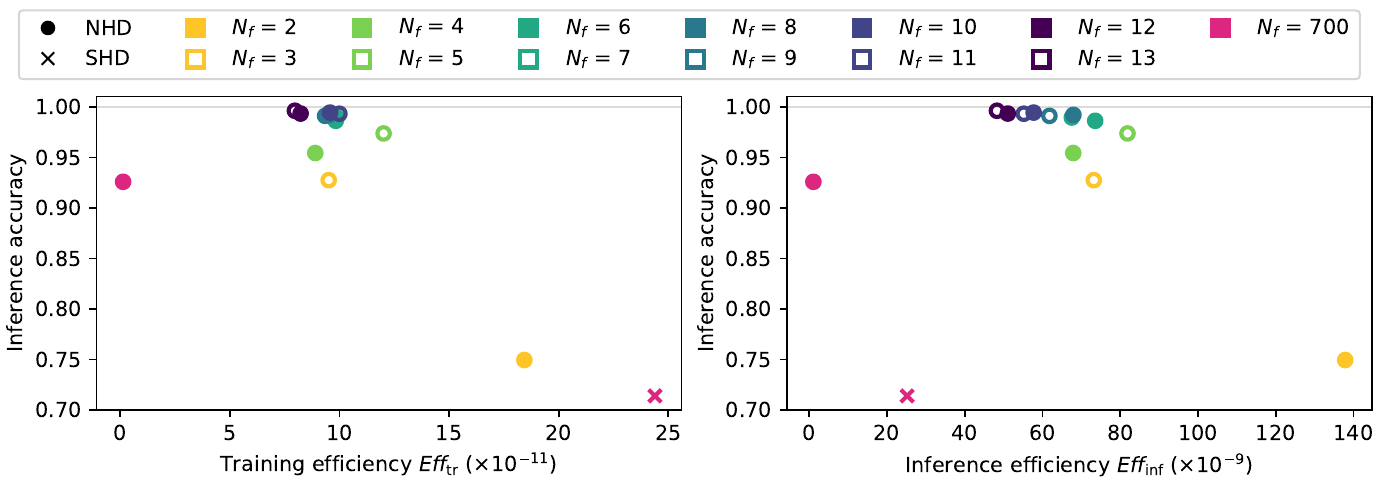}
    \end{subfigure} \\
    \begin{subfigure}[c]{\textwidth}
        \caption{Comparison between NSC variants and SSC.}
        \label{fig:ComparisonFilterNSC}
        \includegraphics[width=\textwidth]{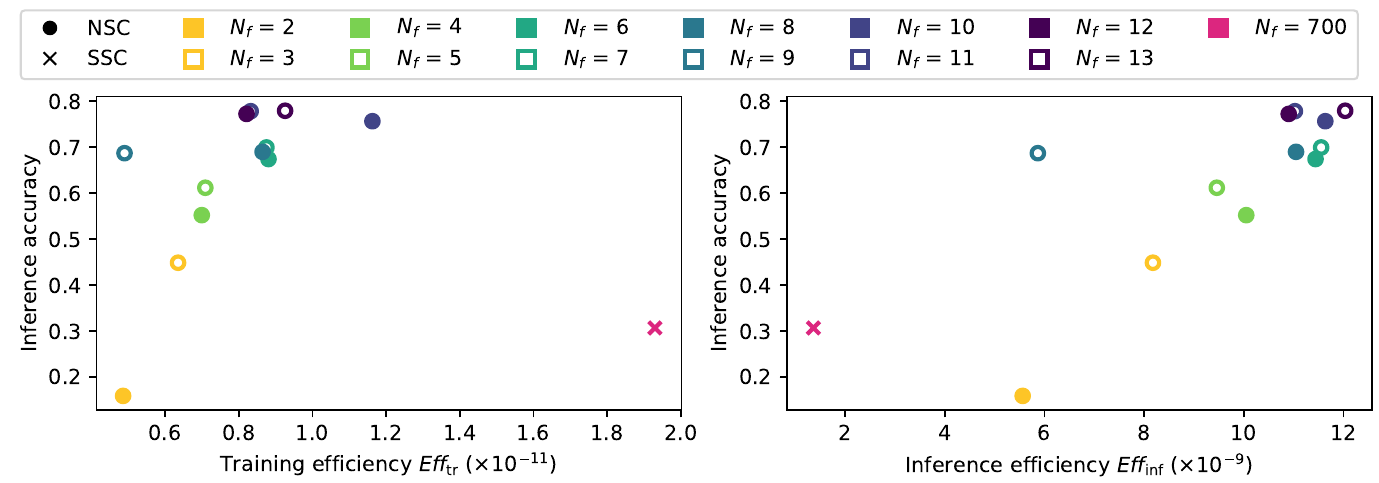}
    \end{subfigure} \\
    \begin{subfigure}[c]{\textwidth}
        \caption{Comparison between NTIMIT variants.}
        \label{fig:ComparisonFilterNTIMIT}
        \includegraphics[width=\textwidth]{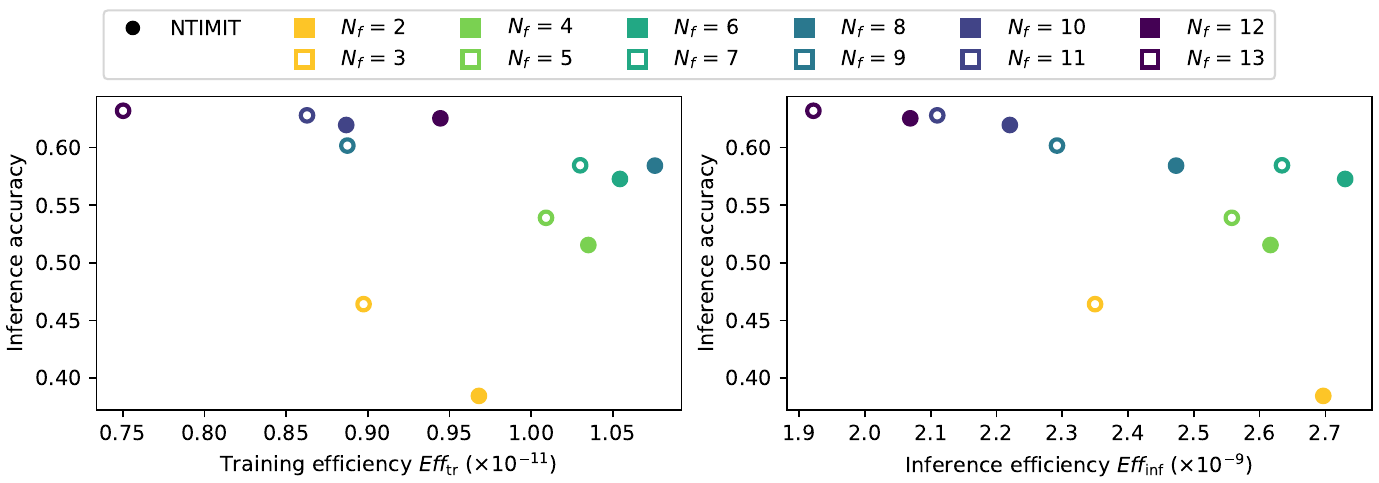}
    \end{subfigure} \\
    \caption{Impact of the number of encoding filters $\Nf$ on classification accuracy and efficiency. HD (a), GSC (b) and TIMIT (c) are encoded with our HLP encoder (round markers) using $\Nf$ filters varying from 2 to 13 (and 700 for HD only). The inference accuracy (y-axis), training efficiency (left column, x-axis) and inference efficiency (right column, x-axis) are obtained using the classifier architecture $\Nf$-300-300-$\No$, each averaged over 3 runs. The accuracy corresponds to the best value obtained over 200, 400 or 100 epochs for HD, SSC and TIMIT respectively. Pink crosses mark the performance of our simple SNN classifier on benchmark datasets SHD and SSC. Respective $\No$ values can be found in Tab.~\ref{tab:datasets} for each dataset. To be noted that the differences between training and inference efficiencies are discussed in Sec.~\ref{sec:diff_eff}.}
    \label{fig:ComparisonFilter}
\end{figure*}

\begin{figure*}[ht]
    \centering
    \includegraphics[width=\textwidth]{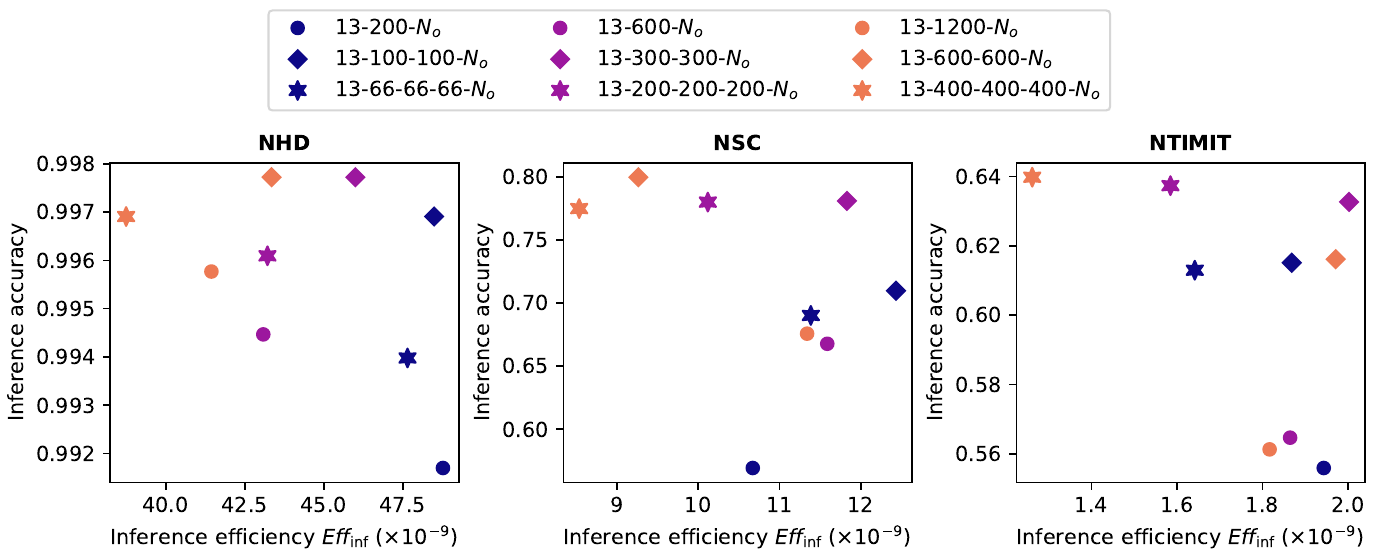}
    \caption{
    Impact of our linear classifier's architecture on classification performance of NHD (left), NSC (middle) and NTIMIT (right), in terms of inference accuracy (x-axis) and efficiency (y-axis). Three network depths are evaluated: either one (round markers), two (diamond) or three (star) hidden layers. Conjointly, three network capacity levels are evaluated: either 200 (blue markers), 600 (purple) or 1200 (orange) LIF neurons uniformly compose the network's hidden layers. Respective $\No$ values can be found in Tab.~\ref{tab:datasets} for each dataset. All encoding parameters are set according to Tab.~\ref{tab:HLP_encoder_param}. The performances are the best ones obtained over 400 epochs (100 for NTIMIT) after averaging over 3 runs.
    }
    \label{fig:ComparisonArchi}
\end{figure*}

\subsubsection{Optimisation of filter distribution}

In our precedent work~\cite{meunier_aicas2025}, we analysed on a smaller scale the impact of the number of filters \Nf involved in encoding, comparing a biomimetic approach where $\Nf = 700$ (similar to Lauscher~\cite{SHD}), with a bio-inspired approach using $\Nf = 13$.
The value of the latter approach comes from the 13 coefficients often used within the standard Mel-Frequency Cepstral Coefficients (MFCCs) method~\cite{MFCC} for speech features extraction.
This section studies the number of filters \Nf but also their spectral placement to optimise the accuracy and efficiency of the pipeline. 

The logarithmic spacing of central frequencies is inspired by the biological cochlear tonotopy, and is empirically superior to linear spacing for speech features~\cite{glasberg1990derivation,slaney1998auditory}. The filter centres are thus placed on a logarithmic grid between the lowest and highest band $F_{\max}$. Changing \Nf or $F_{\max}$ impacts the spacing between each filter.
Fig.~\ref{fig:ComparisonFilter} presents the evolution of classification accuracy (ordinates) and efficiency (abscissa) according to \Nf varying from 2 to 13 for HD, GSC and TIMIT (respectively a, b and c), plus $\Nf = 700$ for HD.
The column on the left uses the training efficiency, while the right one uses the inference efficiency. These accuracies were obtained using a SNN classifier with two hidden layers of 300 units each. 

Fig.~\ref{fig:ComparisonFilterNHD} confirms the results presented in~\cite{meunier_aicas2025}: the NHD variants outperform SHD in terms of accuracy and training efficiency, even with far fewer filters. In particular, using $\Nf=700$ (as in the Lauscher encoder~\cite{SHD}) offers no benefit in our setting: it raises energy and resource usage while yielding lower accuracy than leaner configurations. Using fewer filters significantly reduces the number of input neurons thus the number of synaptic events within the classifier, leading to proportional savings in memory traffic and energy~\cite{ENERGYDampfhoffer,NbrOperationDampfhoffer}. It also eases hardware implementation by lowering resource requirements for the whole pipeline. Consequently, we do not consider $\Nf=700$ further and focus on $\Nf\le 13$. 
According to Fig.~\ref{fig:ComparisonFilterNSC}, the NSC variants surpass the SSC in terms of accuracy and training efficiency once $\Nf \ge 3$. The NSC variant encoded with $\Nf=7$ slightly outperforms those with $\Nf = 8$ or  $\Nf = 9$: filter placement seems to matter as much as the number of filters. As for NTIMIT, we can conclude from Fig.~\ref{fig:ComparisonFilterNTIMIT} its accuracy seems to reach a plateau around $\Nf = 10$, similarly to NHD and NSC.

For all $F_{\max}$ settings and datasets we tested, accuracy saturates at $\Nf \approx 13$. To be conservative and robust across datasets, we adopt the number of filters $\Nf=13$ as default, relaxing to smaller \Nf only when explicitly trading a small accuracy loss for additional efficiency. Section~\ref{sec:discussions} further analyses the impact of $F_{\max}$ on filter placement and performance.

\subsection{Selection of the SNN architecture}
 
The objective of this section is to identify the most suitable SNN architecture \textit{via} which to classify encoded datasets. We evaluate networks one to three layers deep, while varying among three capacity levels, i.e. three network widths. The units are distributed uniformly across hidden layers to keep the total number of hidden units fixed, thus allowing us to study how to best allocate a given compute budget.

As shown in Fig.~\ref{fig:ComparisonArchi}, the highest accuracies were obtained with networks with two hidden layers for NHD (with $\Nh=300$ in each hidden layer) and NSC (with $\Nh=600$ in each hidden layer), and three hidden layers for TIMIT (with $\Nh=400$ in each hidden layer). The complexity of the best performing architecture rises with the difficulty of the classification task, the latter also affecting the classification performance: our simple SNN classifier reaches a near perfect accuracy of $99.77\%$ on a variant of the clean and ``easy'' NHD dataset; however, it only reaches $79.96\%$ and $63.98\%$ for NSC variant and NTIMIT variant respectively, due to the complexity, higher number of labels and inter-sample variation of these two datasets (as previously discussed in Sec.~\ref{sec:datasets}).

Overall, the architecture $\Nf-300-300-\No$ delivers a strong trade-off between accuracy and efficiency across datasets: when it does not allow for the best accuracy, it stays within $2\%$ of the best performing configuration while achieving a similar (NTIMIT) or higher (NHD and NSC) efficiency than the best accuracy configurations.
Moreover, this architecture exhibits stable learning across all studied datasets, without any abrupt drops in test accuracy nor signs of overfitting, even over long runs.
Consequently, unless otherwise stated, the SNN classifiers used in the remainder of this work will adopt this architecture: two hidden layers architecture with $\Nh=300$ units per hidden layer.

\begin{figure*}[ht]
    \centering
    \includegraphics[width=\textwidth]{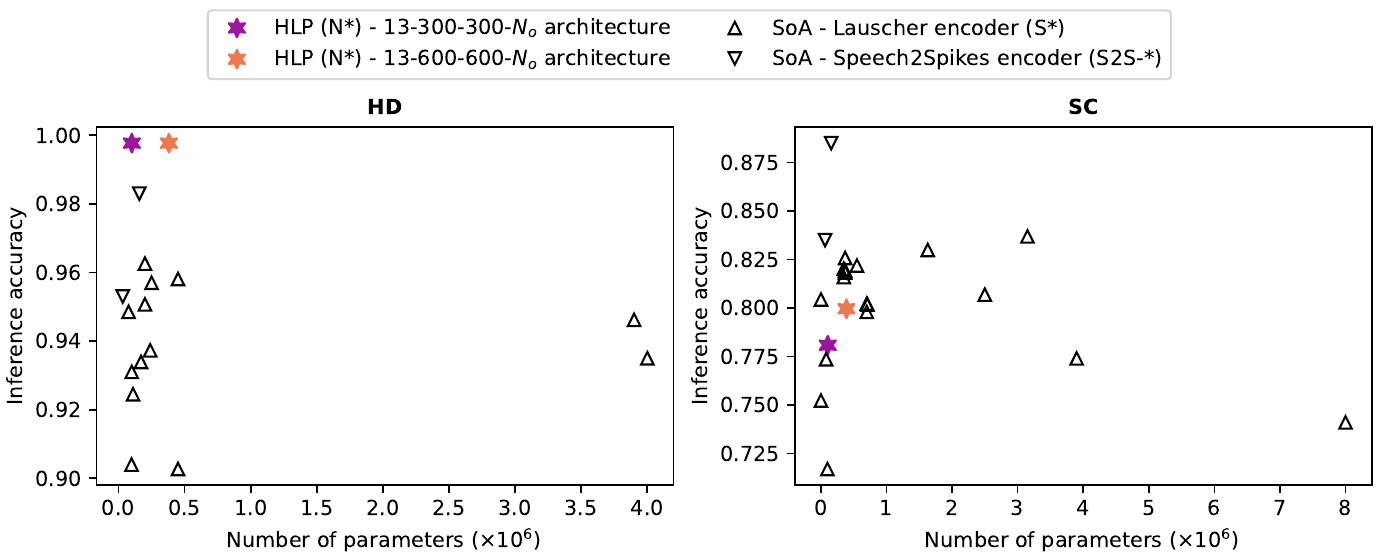}
    \caption{
    Comparison of recent state-of-the-art approaches to spiked-encoded HD (left) and GSC (right) classification. This selection solely encompasses end-to-end neuromorphic pipelines, composed of a bio-inspired, non-learnable cochlear encoder and an SNN classifier (see Sec.~\ref{subsec:scope}). Three encoders are thus selected: the Lauscher encoder (SHD and SSC, upwards triangle markers)~\cite{sun2025pfa,baronig2024advancing,bittar2022surrogate,hammouamri2024dcls,Nowotny2025,sun2023adaptive,queant2025delreclearningdelaysrecurrent,deckers2024co,patino2023empirical,wang2024spikescr,fabre2026,xu2025}; the Speech2Spikes encoder (S2S-HD and S2S-GSC, downwards triangle markers)~\cite{speech2spikes,richter2025}; and our HLP encoder (NHD and NSC, star markers). The number of parameters (x-axis) is used as a convenient proxy for network complexity, thus theoretical energy use. Our results are obtained using a 13-300-300-$\No$ (purple marker) or 13-600-600-$\No$ (orange) architecture, corresponding to the best ones obtained over 400 epochs after averaging over 3 runs.
    }
    \label{fig:plotcross}
\end{figure*}

\begin{figure}[b!]
    \centering
    \includegraphics[width=\columnwidth]{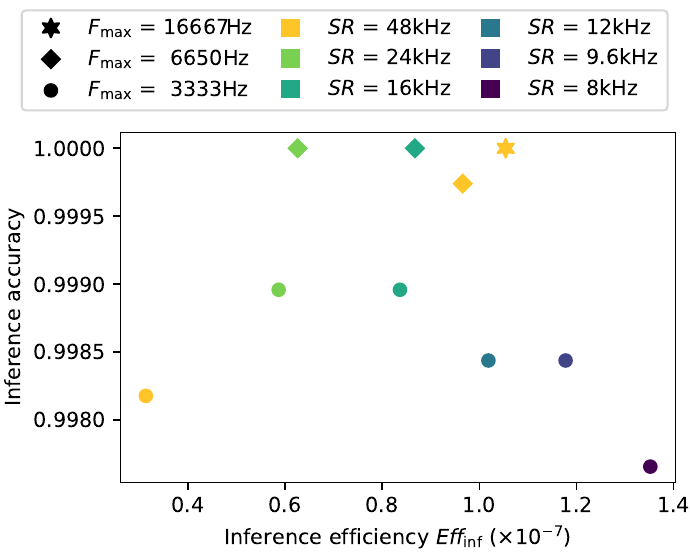}
    \caption{Impact of higher band $F_{\max}$ setting and audio downsampling on downstream classification of NHD-English (subset of NHD comprising only English-spoken digits). The central frequency of the encoder's highest filter $F_{\max}$ is decreased from \qty{16667}{\hertz} (star marker) to \qty{6650}{\hertz} (diamond) and \qty{3333}{\hertz} (round). When possible, the sampling rate $SR$ is reduced from \qty{48}{\kilo\hertz} to \qty{24}{\kilo\hertz}, \qty{16}{\kilo\hertz}, \qty{12}{\kilo\hertz}, \qty{9.6}{\kilo\hertz} or \qty{8}{\kilo\hertz} (respectively downsampled by a factor 2, 3, 4, 5 or 6 before HLP encoding). All other encoding parameters are set according to Tab.~\ref{tab:HLP_encoder_param}. Performances correspond to the best ones obtained with a 13-100-100-10 architecture over 200 epochs, after averaging over 5 runs.
    }
    \label{fig:ComparisonFmax}
\end{figure}

\subsection{Comparison with the state-of-the-art}

As there is no widely accepted, easy to evaluate during early design and platform-agnostic metric for reporting energy consumption in neuromorphic computing, direct comparison within the literature is complex and relevant reports are scarce. As a pragmatic proxy, the number of parameters of a SNN model, i.e. the sum of weight and bias variables, is commonly used as it correlates with model size and memory traffic.  In this section, we compare the results obtained on NHD and NSC using our simple neuromorphic pipeline with SNN classification results obtained on HD and GSC encoded into spikes \textit{via} a non-learnable encoder (such as Lauscher~\cite{SHD} or Speech2Spikes~\cite{speech2spikes}). Fig.~\ref{fig:plotcross} compares together the performance obtained on NHD, on SHD~\cite{sun2025pfa,baronig2024advancing,bittar2022surrogate,hammouamri2024dcls,Nowotny2025,sun2023adaptive,queant2025delreclearningdelaysrecurrent,deckers2024co,patino2023empirical} and on S2S-HD~\cite{speech2spikes,richter2025}; similarly, it compares the accuracies obtained on NSC, SSC~\cite{sun2025pfa,bittar2022surrogate,hammouamri2024dcls,baronig2024advancing,Nowotny2025,queant2025delreclearningdelaysrecurrent,fabre2026,xu2025,deckers2024co,wang2024spikescr} and on S2S-GSC~\cite{speech2spikes,richter2025}. It demonstrates that our simple neuromorphic pipeline attains higher accuracy with fewer parameters than similar pipelines for the spiked encoded HD. Given the relative simplicity of our HLP encoder and SNN classifier's architecture, this suggests that our pipelines have a lower energy requirement for a better performance on HD. Fig.~\ref{fig:plotcross} also shows that the state-of-the-art outperforms our pipeline for spiked encoded GSC -- part of Sec.~\ref{sec:discussions} is dedicated to investigating the causes of this behaviour.

A slightly more informative structural proxy would be the number of synapses of the SNN classifier, since synaptic transmissions are tightly linked to the dynamic energy cost of the neural network. As this dynamic cost is proportional to the number of synaptic operations, even a naive reduction in synapse count typically lowers energy. However, since the efficiency metrics we introduced earlier keep track of both performance accuracy and number of synaptic operations involved in this performance, they will be used to more thoroughly evaluate the energy cost of our pipelines in the remainder of this paper.

\subsection{Optimising encoding for efficiency}
\label{subsec:filtdistrib}

\begin{figure*}[!t]
    \centering
    \includegraphics[width=\textwidth]{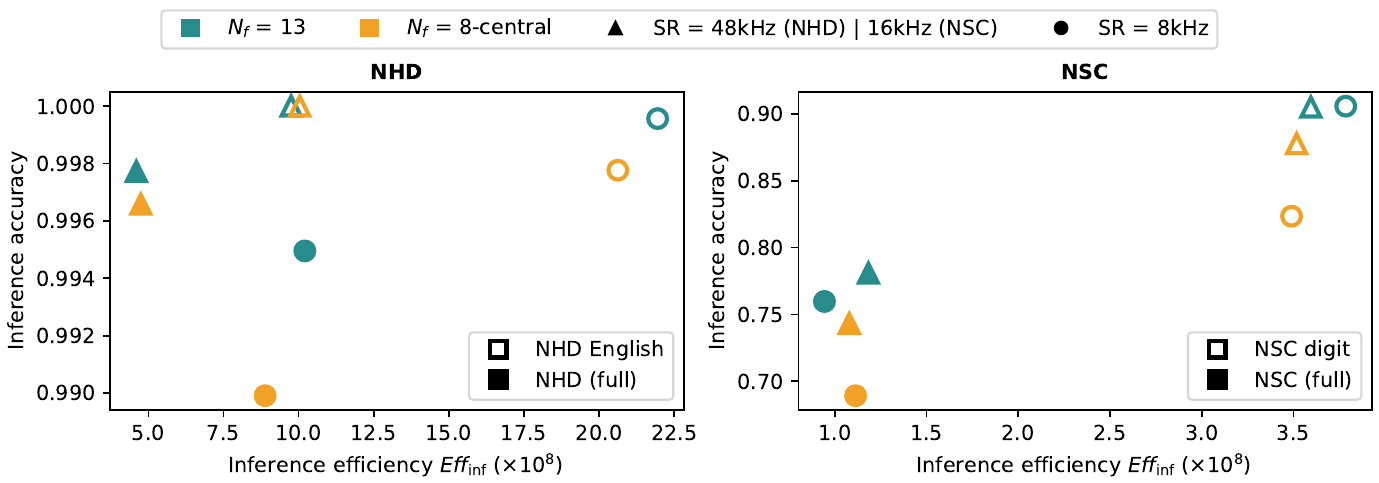}
    \caption{
    Impact of audio down-sampling and filter clipping on downstream classification of NHD (left plot), NSC (right plot) and their subsets. NHD is pictured with NHD-English, its subset comprising only English-spoken digits; NSC is pictured with NSC-digit, its subset with only samples corresponding to digits. The classification accuracy and efficiency at inference is evaluated over variants of each dataset and subset: either with all 13 original encoding filters (turquoise markers) or only the 8 central filters after clipping (orange markers); and either with original sampling rate (triangle markers) or after audio down-sampling to \qty{8}{\kilo\hertz} (round markers). All other encoding parameters are set according to Tab.~\ref{tab:HLP_encoder_param}. The performances correspond to the best ones obtained with a $\Nf$-300-300-$\No$ architecture (where $\Nf$ is either 13 or 8 after filter clipping, and $\No$ depends on the dataset used) over 400 epochs after averaging over 3 runs.
    }
    \label{fig:ComparisonEncoding}
\end{figure*}

\subsubsection{Filters' sampling rate and frequency}

This section aims to raise efficiency while maintaining high accuracy using two strategies: down-sampling and clipping filters.
As audio sampling rates vary across datasets, we study the impact on performance of down-sampling raw audio prior to spike encoding, as well as lowering the highest filter centre frequency $F_{\max}$ according to the Nyquist limit.
Indeed, since down-sampling (i.e. reducing the sampling rate) reduces the value of the Nyquist frequency, the encoder's top band must hence be lowered or clipped to keep all pass-bands within the representable frequency range. 

We use in this study a smaller SNN architecture ($\Nf-100-100-\No$) to magnify the impact of down-sampling and lowering the filter-bank ceiling on downstream performance. Fig.~\ref{fig:ComparisonFmax} compares classification results obtained on a subset of NHD, comprising only the English-spoken digits (thus reducing the number of labels \No from 20 to 10). This subset is either presented at its original sampling rate of \qty{48}{\kilo\hertz}, or down-sampled to a sampling rate comprised within \{24; 16; 12; 9.6; 8\}~\si{\kilo\hertz}. When feasible, to decouple the effect of sampling rate from the filter-bank ceiling, variants of this subset are generated for a $F_{\max}$ lowered to \qty{6.650}{\kilo\hertz} or \qty{3.333}{\kilo\hertz} from its original value \qty{16.667}{\kilo\hertz}.

Fig.~\ref{fig:ComparisonFmax} shows that classification accuracies remain within one thousandth of each other when down-sampling the English subset of NHD down to \qty{8}{\kilo\hertz}, whether using the most adequate or a lowered $F_{\max}$. A similar study was led on a subset of NSC comprising only spoken digits: it compared its original 13 filters distribution and a novel NSC encoding using the 13 filters used for NHD, minus the two upper filters -- clipped as they reach above GSC's Nyquist frequency (since GSC and HD have different sampling rates).
The results remain quite similar: the performance accuracies stay within one thousandth of each other and learning curves also exhibit similar convergence speed across all encoded variants. 

As such, we validate setting $F_{\max}$ to \qty{16.667}{\kilo\hertz} for data with a sampling rate of \qty{48}{\kilo\hertz}, to \qty{6.65}{\kilo\hertz} for a sampling rate of \qty{16}{\kilo\hertz} and to \qty{3.333}{\kilo\hertz} for a sampling rate of \qty{8}{\kilo\hertz}. \\

\subsubsection{Filters' distribution}
Moreover, as portrayed in Fig.~\ref{fig:encoder_wide}~D), a logarithmic distribution of $\Nf=13$ filters leads to less active filters in the extremas: due to the distribution of the audio frequencies within the raw samples, the three lower and two upper filters fire markedly fewer spikes than the rest when using the highest $F_{\max}$. A lower $F_{\max}$ would reduce the gap of spikes count between these five filters and the other central filters -- however, removing them altogether might not impact the final accuracy while significantly improving the efficiency (due to a reduced number of input channels). We thus assess in this section the importance of these five filters for the classification performance and efficiency of our pipelines according to the complexity and the sampling rate of the datasets.
A simple pruning strategy is adopted to this end: the same logarithmic grid is kept, and the aforementioned filters are clipped before encoding, retaining only the eight central filters. If successful, this strategy would reduce the number of neurons, synapses and synaptic events, easing hardware implementation and slightly lowering energy cost.

Fig.~\ref{fig:ComparisonEncoding} presents the results of applying this clipping strategy combined with down-sampling to \qty{8}{\kilo\hertz} the datasets NHD and its English subset, as well as NSC and its digit subset. Both NHD-English subset and NSC-digit subset are studied here as an intermediate, easier step towards more difficult tasks. According to the results, these strategies improve the downstream efficiency by up to a factor 2 while keeping an accuracy within one thousandth of the baseline dataset. For ``easy'' tasks (i.e. NHD-English subset and NHD -- see Sec.~\ref{sec:datasets}), the classification accuracy is little affected by the clipping of the encoding filters. Similarly, the audio dataset can be down-sampled to \qty{8}{\kilo\hertz} without significant impact on the accuracy.
However, these strategies did not lead to consistent gains of efficiency across the datasets. For the more challenging datasets (NSC-digit subset and NSC), clipping and down-sampling remove critical data for the learning phase. Clipping low-activity filters and pre-encoding down-sampling are not always effective strategies to reduce event rates and network activity. Their benefit is dataset-dependent: highly learnable tasks tolerate both strategies well, whereas harder tasks require the full filter bank and higher bandwidth to maintain accuracy.

\section{Discussions}
\label{sec:discussions}

\subsection{Assessing the differences in NHD and NSC performance}
\label{subsec:compHDSC}

To identify the reason behind the significant differences between NHD and NSC results yielded by our simple pipeline, we align the classification tasks by restricting NHD to its NHD-English subset (i.e. only English-spoken digits) and NSC to its NSC-digit subset (i.e. only samples corresponding to digits).
We probe the dissimilarity issue by evaluating NHD-English and NSC-digit with a simple SNN classifier of architecture $\Nf-100-100-\No$. To be noted that $\No=10$ for NHD-English and NSC-digit as the subsets' labels are both reduced to the 10 digits.
In addition, both NHD-English and NSC-digit are encoded without RMS scaling to even out the comparison. 

As seen in Tab.~\ref{tab:subsets_results}, these supposedly similar subsets yield markedly different outcomes: NHD-English scores 12.7 points more than NSC-digit. Since the pipelines are matched and the datasets' content is similar, these gaps point to dataset-specific factors -- more specifically the number of speakers involved. Indeed, the GSC samples corresponding to digits were recorded from 2519 different speakers, unequally represented (75\% of the speakers are represented at most 13 times each); whereas HD is recorded from only 12 speakers, represented between 470 to 1050 times each. To further evaluate this hypothesis, we assess a second subset of NSC-digit, comprising all samples uttered by the 12 most represented speakers. This ``NSC-digit -- top 12'' subset contains 66 to 112 samples per speaker, and leads to a classification accuracy of 99.22\%, narrowing the gap with NHD-English. The lower performance on NSC-digit, and more globally on NSC might largely stem from greater inter-speaker variability and differences in speaker representation. To ensure that this gap is not an artefact of uneven sample distribution, we perform a final comparison with a second, size-matched subset of NHD-English: this ``NHD-English -- 20\%'' contains a random selection of 20\% of the NHD-English dataset samples, uniformed over label, i.e. approximately the same number of samples as ``NSC-digit -- top 12''; and reaches a similar accuracy, confirming that dataset size alone does not explain the discrepancy.

\begin{table}[h]
    \centering
    \caption{
    Impact of varying sample selection on the classification of NHD, its English subsets, NSC and its digit subsets. The accuracies are obtained with a SNN classifier of architecture $\Nf$-100-100-$\No$, run over 400 epochs after averaging over 3 runs.} 
    \begin{tabular}{l c c c c}
        \hline
        Data subset & \# samples & \# speakers & Acc. & $\EffInfMetric$ \\
        \hline
        NHD & \qty{10420}{\nothing} & 12 & \qty{99.69}{\percent} & \qty{48.49}{\micro\nothing} \\
        NHD-English & \qty{5090}{\nothing} & 12 & \qty{99.85}{\percent} & \qty{110.75}{\micro\nothing} \\
        NHD-English -- 20\% & \qty{1030}{\nothing} & 12 & \qty{100}{\percent} & \qty{184.19}{\micro\nothing} \\
        \hline
        NSC & \qty{105830}{\nothing} & \qty{2618}{\nothing} & \qty{70.97}{\percent} & \qty{12.43}{\micro\nothing} \\
        NSC-digit & \qty{38908}{} & \qty{2519}{\nothing} & \qty{87.61}{\percent} & \qty{27.01}{\micro\nothing} \\
        NSC-digit -- top 12 & \qty{1032}{\nothing} & 12 & \qty{99.22}{\percent} & \qty{610.18}{\micro\nothing} \\
        \hline
    \end{tabular}
    \label{tab:subsets_results}
\end{table}

\subsection{Exploring the training and inference efficiencies}
\label{sec:diff_eff}

\begin{figure*}[ht]
    \centering
    \includegraphics[width=\textwidth]{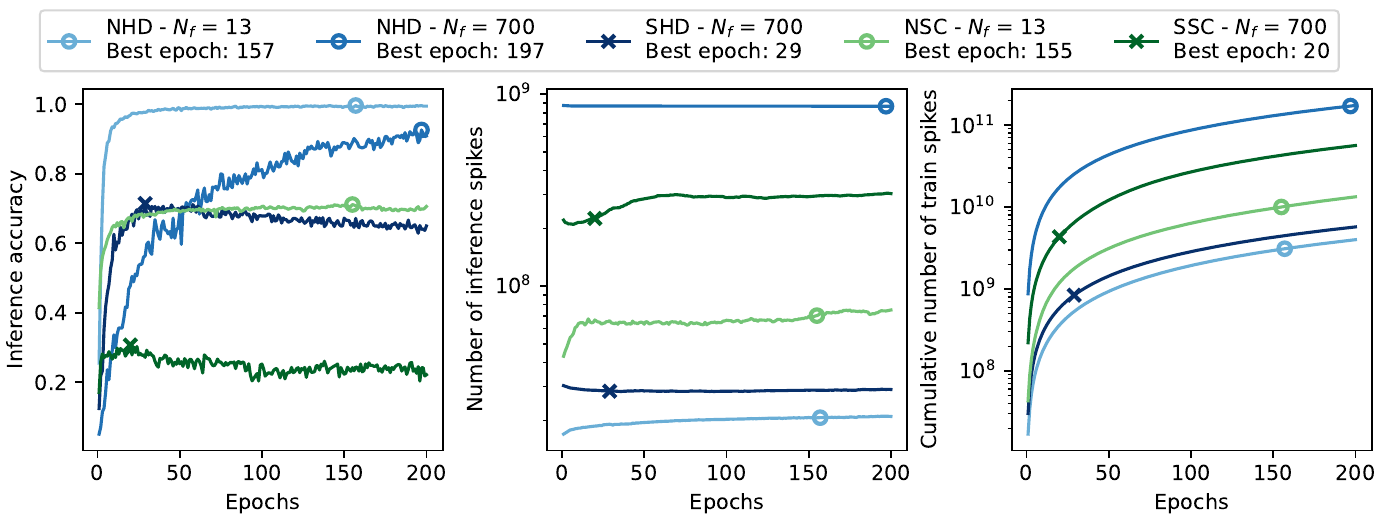}
    \caption{Evolution of accuracy and number of spikes propagated within a SNN classifier of architecture $\Nf$-300-300-$\No$ over 200 epochs, averaged on 3 runs. The inference accuracy (left plot), number of spikes propagated during inference at one epoch (middle) and cumulative number of spikes propagated during training over the previous epochs (right) are illustrated for NHD encoded with $\Nf=13$ (light blue) and $\Nf=700$ (blue), for NSC encoded with $\Nf=13$ (light green) as well as for SHD (dark blue) and SSC (dark green). The markers (round for HLP encoding and cross for Lauscher encoding) indicate the best epoch, i.e. the epoch at which the inference accuracy is at its highest for each dataset (as indicated in the legend).}
    \label{fig:efficiencies}
\end{figure*}

We introduced in Sec.~\ref{subsec:metrics} two efficiency metrics, assessing the trade-off between classification accuracy and number of spikes propagated within the network. The goal of this work is to simultaneously increase the accuracy and reduce the number of spikes, correlated to the theoretical energy efficiency -- we are therefore targeting a high efficiency. We defined two variants of efficiency: one assessing the energy cost of training to reach a certain accuracy (training efficiency \EffTrMetric), the other assessing the cost of inferring an accuracy at its corresponding epoch (inference efficiency \EffInfMetric). We evaluated both metrics in Fig.~\ref{fig:ComparisonFilter}: in most cases, among data variants with a varying $\Nf$, the dynamics of $\EffTrMetric$ and $\EffInfMetric$ are quite similar. Two main outliers emerge: the case of NSC encoded with $\Nf = 2$, where its $\EffTrMetric$ is on the right side of the larger group whereas its $\EffInfMetric$ is on the left; and the similar case of both Lauscher-encoded datasets SHD and SSC. 
We investigate the latter in Fig.~\ref{fig:efficiencies}, which illustrates the accuracy and the number of propagated spikes of the \mbox{\Nf{}{\kern -0.7ex}-300-300-\No} network used in Fig.~\ref{fig:ComparisonFilter}, over 200 epochs. Fig.~\ref{fig:efficiencies} (left) identifies the best accuracy as reported in this work, i.e. the highest inference accuracy over the total training epochs.
The corresponding best epoch is reported in the legend. It is noteworthy that the Lauscher-encoded SHD and SSC achieve their best performance with our simple SNN classifier at an early epoch (29$|$20 out of 200), whereas the HLP-encoded NHD and NSC have to learn over at least 150 epochs to reach their highest point -- even though the gain is little compared to the accuracy obtained at epoch 29$|$20 for example. This does not mean that HLP encoding requires a learning phase slower than Lauscher: both NHD and NSC encoded with $\Nf=13$ reach a plateau as soon as, or even sooner than, SHD and SSC. Additionally, this plateau is maintained for the HLP-encoded datasets whereas their Lauscher counterparts show clear overtraining behaviour, i.e. an accuracy decrease over a long learning phase, after reaching their peak.

This first observation regarding the various datasets' best epochs brings out an answer to the discrepancies identified in Fig.~\ref{fig:ComparisonFilter}: as SHD and SSC make use of a small learning phase, their training efficiency $\EffTrMetric$ highly benefits from the low cumulative number of train spikes. 
Fig.~\ref{fig:efficiencies} (middle) and (right) confirm our theory by respectively displaying the number of spikes propagated at inference at each epoch, and accumulated over the training phases. The number of spikes at inference remains quite stable over the epochs, with a surprising increase combined with a reduced accuracy for the SSC dataset. The length of the learning phase has therefore low impact on the inference efficiency \EffInfMetric, marking it as the most reliable. The cumulative number of spikes over training however increases by definition at each epoch. The training efficiency $\EffTrMetric$ thus strongly depends on the learning dynamics over the epochs, as even one additional learning phase increases linearly the divisor of this ratio. It additionally challenges the traditional choice of the neuromorphic AI community to report their performance with the best accuracy, and not the last accuracy achieved -- as to obtain this ``best value'', one must still perform learning and inference over all the epochs to identify their peak. 

Further studies reveal that the NSC variant encoded with $\Nf=2$ (see Fig.~\ref{fig:ComparisonFilter}) also benefits from such unexpected bias of \EffTrMetric: its low accuracy (10.61\%) is effectively balanced out by a significantly early learning peak (at epoch 8 out 200), thus leading to a high training efficiency but quite small inference one. All in all, these observations lead us to give more importance to and solely report the inference efficiency $\EffInfMetric$ in experiments following the one described in Fig.~\ref{fig:ComparisonFilter}. 

To be noted that the NHD variant encoded with $\Nf=700$ does require a slower learning phase than all others evaluated in Fig.~\ref{fig:efficiencies}. The best epoch is quite close to the end of the learning phase (197 over 200) and the overall tendency of the curve seems to indicate that it does not reach an accuracy plateau before the last epoch. This tends to show that a higher number of encoding channels leads to a longer learning phase, thus further promotes our choice of a low $\Nf$ as opposed to Lauscher's 700 channels.

\subsection{Comparison with state-of-the-art models}

\begin{table*}[ht]
    \centering
    \caption{Number of parameters and inference performance of state-of-the-art neuromorphic approaches to audio classification, combining non-learnable cochlear encoder and SNN classifier.}
    \label{tab:sotapipeline}
    \small
    \setlength{\tabcolsep}{4pt}
    \renewcommand{\arraystretch}{1.15}
    \begin{tabularx}{\textwidth}{@{} l l >{\raggedright\arraybackslash}X c c c @{}}
        \toprule
        Audio & Cochlear encoder & SNN classifier & Param. & Acc. (\%) & Ref. \\
        \midrule
        \multicolumn{5}{@{}l}{\emph{HD pipelines}} \\
        SHD & Lauscher & ffSNN with d-cAdLIF (2 HLs) & \qty{76}{\kilo\nothing} & 94.85$\pm$0.64 & \cite{deckers2024co} \\
        SHD & Lauscher & ffSNN with DCLS for delay learning (2 HLs and 1 KC) & \qty{200}{\kilo\nothing} & 95.07$\pm$0.24 & \cite{hammouamri2024dcls} \\
        SHD & Lauscher & SpikeSCR: fSNN with KDCL (1 global-local encoder, 8 attention heads and hidden size 128) & \qty{250}{\kilo\nothing} & 95.7 & \cite{wang2024spikescr} \\
        SHD & Lauscher & rSNN with Symplectic-Euler adLIF & \qty{450}{\kilo\nothing} & 95.81 & \cite{baronig2025adlifsnn} \\
        SHD & Lauscher & ffSNN with parameter-free attention & \qty{200}{\kilo\nothing} & 96.26 & \cite{sun2025pfa} \\
        S2S-HD & Speech2Spikes & lSNN (40--120--120--20) & \qty{32}{\kilo\nothing} & 95.3 & \cite{richter2025} \\
        S2S-HD & Speech2Spikes & lSNN (80--256--256--256--20) & \qty{157}{\kilo\nothing} & 98.3$\pm$0.3 & \cite{speech2spikes} \\
        \textbf{NHD} & \textbf{HLP (ours)} & \textbf{lSNN (13--100--100--20)} & \textbf{\qty{13.5}{\kilo\nothing}}  & \textbf{99.69} & \textbf{(this work)} \\
        \textbf{NHD} & \textbf{HLP (ours)} & \textbf{lSNN (13--300--300--20)} & \textbf{\qty{100.5}{\kilo\nothing}}  & \textbf{99.77} & \textbf{(this work)} \\
        \midrule
        \multicolumn{5}{@{}l}{\emph{GSC pipelines}} \\
        SSC & Lauscher & ffSNN with DCLS for delay learning (2 HLs and 1 KC) & \qty{700}{\kilo\nothing} & 79.77$\pm$0.09 & \cite{hammouamri2024dcls} \\
        SSC & Lauscher & ffSNN with DCLS for delay learning (3 HLs and 2 KCs) & \qty{2.5}{\mega\nothing} & 80.69$\pm$0.21 & \cite{hammouamri2024dcls} \\
        SSC & Lauscher & rSNN with adaptive skip recurrent connection (3 HLs) & \qty{370}{\kilo\nothing} & 81.93 & \cite{xu2025asrcsnn} \\
        SSC & Lauscher & ffSNN with SSM-inspired LIF (2 HLs) & \qty{350}{\kilo\nothing} & 82.03$\pm$0.25 & \cite{fabre2026} \\
        SSC & Lauscher & DelRec: rSNN with delay learning in recurrent connections (3 HLs) & \qty{370}{\kilo\nothing} & 82.58$\pm$0.08 & \cite{queant2025delreclearningdelaysrecurrent} \\
        SSC & Lauscher & SpikeSCR: fSNN with KDCL (2 global-local encoders, 16 attention heads and hidden size 256) & \qty{3.15}{\mega\nothing} & 83.69 & \cite{wang2024spikescr} \\
        S2S-GSC & Speech2Spikes & lSNN (40--120--120--35) & \qty{64}{\kilo\nothing} & 83.5 & \cite{richter2025} \\
        S2S-GSC & Speech2Spikes & lSNN (80--256--256--256--35) & \qty{157}{\kilo\nothing} & 88.5$\pm$0.1 & \cite{speech2spikes} \\
        \textbf{NSC} & \textbf{HLP (ours)} & \textbf{lSNN (13--600--600--35)} & \textbf{\qty{390}{\kilo\nothing}} & \textbf{79.96} & \textbf{(this work)} \\
        \midrule
        \multicolumn{5}{@{}l}{\emph{TIMIT pipeline}} \\
        \textbf{TIMIT} & \textbf{HLP (ours)} & \textbf{Our SNN (13--400--400--400--39)} & \textbf{\qty{342}{\kilo\nothing}}  & \textbf{63.98} & \textbf{(this work)} \\
        \bottomrule
    \end{tabularx}
    
    \vspace{4pt}
    \footnotesize
    \textit{Note:    ffSNN = feedforward network ; rSNN = recurrent network ; lSNN = linear network ; HL = hidden layer ; \\ d-cAdLIF = constrained adaptive LIF with trainable delays ; DCLS = dilated convolution with learnable spacings ; \\ KC = number of non-zero elements in the kernel ; KDCL = knowledge distillation method based on curriculum learning.}
\end{table*}

The HLP encoder yields denser, more contiguous activity within informative bands and noticeably fewer isolated, noise-like spikes. In contrast, the more bio-mimetic Lauscher encoder produces broader, sparser activity with many singleton events. While the latter is more bio-mimetic, the former allows more efficiency. 

Across pipelines, we observe that our HLP encoder favours a simple linear SNN classifier over more complex architectures, such as the recurrent SNN one used in~\cite{SHD} or those providing state-of-the-art results on SHD and SSC (see Tab.~\ref{tab:sotapipeline}). For example, the classifier from~\cite{hammouamri2024dcls} results in an accuracy of $\sim80\%$ on SSC (effectively reproducing their work) but obtains only $\sim60\%$ on the NSC dataset; conversely, the same model reaches $\sim95\%$ on SHD and a higher $\sim98\%$ on NHD. Meanwhile, our SNN classifier reaches an accuracy of $79.96\%$ on NSC but less than $32\%$ on SSC; and $99.77\%$ on NHD but less than $72\%$ on SHD. 

These results show that our neuromorphic pipeline enables our simple SNN classifier to outperform more complex spiking models on the easier, well-controlled HD dataset. However, on the more heterogeneous GSC dataset, it underperforms pipelines using a more bio-mimetic encoder such as Lauscher paired with a more complex classifier, or a different fixed front-end such as Speech2Spikes with a linear classifier. Put differently, the more bio-mimetic encoders tend to be more resilient to data variability, whereas our bio-inspired HLP encoder is more specialised: it excels when the data distribution is clean and consistent, but it exploits the subtleties of data less effectively.

Indeed, this specialisation of our HLP encoder implies a greater need for homogeneous coverage to remain competitive as variability grows. Consistent with Sec.~\ref{subsec:compHDSC}, restricting GSC to a subset with stronger speaker representation (e.g., the top 12 speakers), the performance gap closes substantially, confirming that our pipelines benefit from sufficient similar data. 

More generally, the high contrasts in performance among different strategies highlight the impact of the choice of encoder on the effectiveness of different SNN classifiers, depending on their structure and complexity. What may appear as ``noisy'' spikes from the more bio-mimetic Lauscher encoder seems to effectively provide informative data, which is not well exploited by simpler classifiers. However, more complex SNN models, able to grasp more complex nuances within the input patterns, achieve better performances thanks to these critical ``noisy'' spikes. 

A way to increase the performances of our neuromorphic pipeline on more complex datasets (i.e. GSC and TIMIT) could be to tune the threshold's value per channel to enhance the spikes produced by quieter channels (with a lower spikes count as is than the other channels) and combine this variability with a more complex downstream classifier, better exploiting the patterns less noticeable among the most obvious ones. We did not explore this path in this work since one of our main goals was to focus on efficiency and ease of hardware implementation. 

Table~\ref{tab:sotapipeline} concludes our simulation study by providing a comparison of our approach with similar neuromorphic pipelines for speech classification. For HD and GSC, we report state-of-the-art works applying SNN models on datasets spiked using Lauscher or Speech2Spikes, alongside our results. In addition, we include our TIMIT pipeline, even though -- to the best of our knowledge -- no other end-to-end neuromorphic pipeline addressing the TIMIT dataset has been published.

\section{Proof of concept in hardware}
\label{sec:hardware}
\subsection{Architecture}

To demonstrate the relevance of this work in the design of hardware neuromorphic smart sensors, we implemented the complete processing pipeline of the \mbox{$\Nf$-100-100-$\No$} architecture used in Tab.~\ref{tab:subsets_results} on the FPGA part of a \Zynq device. 
To keep the architecture footprint small, we only process the English subset of the HD dataset with the eight central filters identified in Sec.~\ref{subsec:filtdistrib} -- we thus set $\Nf = 8$ and $\No = 10$. \\

\paragraph{Differences between software simulation and hardware implementation}
When designing online hardware smart sensor, one key priority is to reduce unnecessary computations, especially when no data is present at the input. 
Thus to demonstrate an architecture compatible with such a future online smart sensor, two main architecture choices were made that differ from previous simulations: 

\begin{itemize}
    \item In our implemented neuromorphic Low-Level Programmable (LLP) encoder, the filters were designed using their Second Order Subsection (SOS) form using fixed point arithmetic and the coefficients were encoded with signed Q9.22 format.  
    However, Infinite Impulse Response (IIR) filters can, after use, present some residual value due to their recursive nature and coefficients quantification, which result in spontaneous activity if the architecture is allowed to run continuously.
    In order to prevent strong spontaneous activity, an arbitrary low threshold was added to the encoding neurons that only integrate values above it.
    This creates an encoding neuron "dead band" for low filter output values.
    
    \item In the implemented SNN blocks, a similar difficulty emerges when targeting an architecture that can run online at low power without spontaneous spikes that generate extra computation.
    Thus, while the implemented architecture is compatible with biases, putting non-zero values can cause spontaneous spikes in the network architecture for neurons with positive biases for a sensor running continuously.
    To prevent this, biases in simulation and in the implemented architecture are set to 0, thus reducing the effective number of parameters by 210.   
\end{itemize} 

Using this architecture, we first encoded the dataset into spikes with the hardware LLP encoder.
The encoded data was then used to run a hardware-aware reference training simulation specific to this implementation to obtain the weights of the SNN before running the complete test from encoding to inference on the \Zynq device. \\

\paragraph{Filters}
The implemented LLP filter properties of the neuromorphic sensor part are shown in Fig.~\ref{fig:encoder_wide}D: the 8 implemented channels appear in dotted lines and overlap the 13 ones used in floating-point simulation, thus highlighting the similarities between simulation and hardware implementation. 
Fig.~\ref{fig:encoder_wide}C pictures the spikes produced by the implemented LLP encoder (in orange) compared to the spikes produced in simulation on the same audio sample (in blue), further emphasising the closeness of our hardware implementation to the previous simulations.  \\

\paragraph{Computation}For each layers, a Finite State Machine (FSM) is triggered for the successive computation of each postsynaptic neuron only when an input spike or discharge event is received from a previous layer.
Furthermore, output spikes are transmitted one by one for each layer in our architecture. 
When a layer is done going through all its output neurons, a discharge event is transmitted to order the next layer to compute its discharge.  \\

\paragraph{Input stream}
The architecture was designed and tested on the English subset of the HD dataset. It can handle an input data stream of a microphone with a \qty{48}{\kilo\hertz} sampling frequency. This ensures that data can be processed within the interval of the sensor sampling speed, thus making it compatible for an online deployment. 
An actual run of this architecture on the whole HD-English subset lead to an average running speed of around \qty{152}{\milli\second} including loading the file, sending the data and retrieving the result for normalised \qty{1.4}{\second} long audio samples done by the ARM processor part with a FPGA part running at \qty{10}{\mega\hertz}.

\begin{figure}[t!]
    \centering
    \includegraphics[width=.8\linewidth]{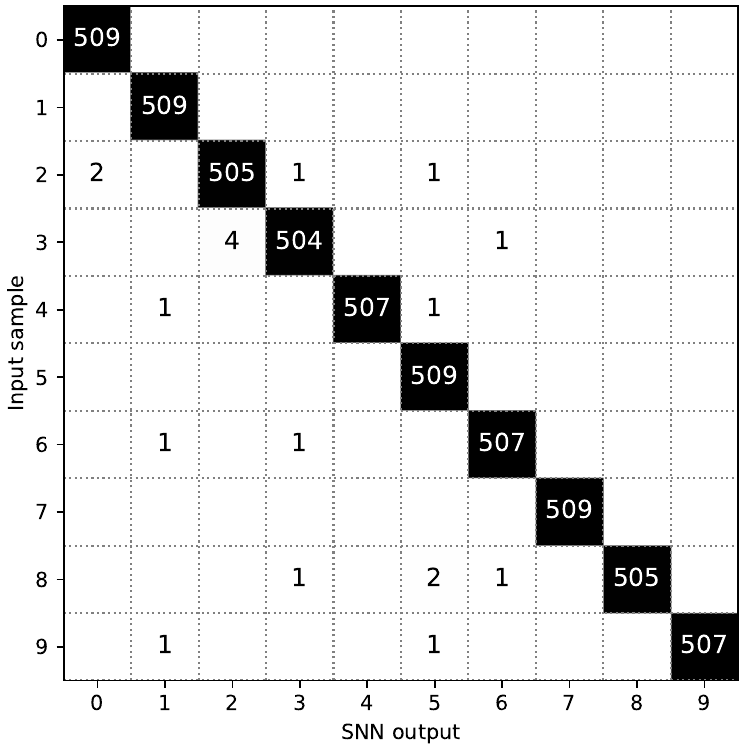}
    \caption{Confusion matrix obtained on HD's English subset with our architecture on FPGA. 
    Of the complete dataset (train and test), only 19 samples were incorrectly detected for a total accuracy rate of \qty{99.63}{\percent}.
    }
    \label{fig:hardware-res}
\end{figure}

\subsection{Results}

Running a reference simulation with hardware-informed constraints over 100 epochs resulted in \qty{99.42}{\percent} train accuracy and \qty{99.21}{\percent} test accuracy.
Using the final weights obtained in this reference simulation, the accuracy obtained by the hardware architecture when presenting the whole dataset (train and test) reaches \qty{99.63}{\percent}. The resulting confusion matrix is shown in Fig.~\ref{fig:hardware-res}.

While those accuracies are lower than previously described simulations, they remain well within the state-of-the art despite the use of a more constrained architecture preventing strong intrinsic spontaneous activity.
This demonstrates that our general processing pipeline makes for a good candidate for online smart sensor deployment. 

Our implementation is quite generic and not fully dedicated to its application, and we intend to further reduce its footprint by removing biases computation states and memories for example. Further works will evaluate the impact of quantification and explore parameters adjustments of our hardware architecture to reach the highest simulation accuracies. 

\section{Conclusion}  

Our work introduced a simple, hardware-friendly, end-to-end neuromorphic pipeline for speech classification converting audio to spikes using a non-learnable, high-level, programmable encoder, and processing these output spikes with a linear SNN classifier.
Our pipeline targets easy, performing and efficient FPGA deployment: the encoder uses IF neurons, sparse events and a small filter bank, all of which reduce memory traffic as well as hardware implementation and parametrisation complexity compared to more bio-mimetic encoders, such as Lauscher~\cite{SHD} or Speech2Spikes~\cite{speech2spikes}.

Our study isolates the encoder’s role in both accuracy and efficiency for the whole pipeline. Using hardware-neutral, spike-count–based metrics, we demonstrated that smaller filter banks are competitive with other approaches. Additionally, simple data reduction \textit{via} down-sampling or filter clipping can raise efficiency by up to a factor 2 while keeping the accuracy within one thousandth on the less complex datasets. Transferring our pipeline to other audio datasets solely requires tuning the encoder's threshold, which controls event density for better efficiency: this hyperparameter can be optimised to reduce energy cost and maintain a high accuracy. Our encoder trades some robustness for specialisation: it excels on well-controlled data but is less tolerant to data variability.
However, despite its limited performance on heterogeneous datasets, our pipeline outperforms state-of-the-art strategies on the Heidelberg Digits dataset.

Finally, our results highlight the necessary conjoint optimisation of encoding and processing spiked audio data for optimising both performance and efficiency. We demonstrate that only a thorough simultaneous exploration of encoder parametrisation (global or channel-wise) and classifier architecture could yield high accuracy combined with low activity.

Future work includes strategies that improve resilience while preserving pipeline simplicity and ease of hardware implementation, narrowing the gap with more bio-mimetic encoders on high-variability datasets. Additionally, we intend to apply our simple pipeline on other datasets and translate spike-count savings into real latency, energy and memory values \textit{via} further FPGA measurements.

\section*{Glossary}
\vspace{0.5cm}

\noindent
\begin{tabular}{c p{6.7cm}}
    \textbf{SNN} & Spiking Neural Network \\
    \textbf{FPGA} & Field-Programmable Gate Arrays \\
    \textbf{HLP} & High-Level Programmable (encoder) \\
    \textbf{LLP} & Low-Level Programmable (encoder) \\
    \textbf{MFCC} & Mel-Frequency Cepstral Coefficient  \\
\end{tabular}

\subsection*{Datasets}
\noindent
\begin{tabular}{>{\centering}m{1.41cm} m{6.6cm}}
    \hline
    \textbf{HD} & Heidelberg Digits (from~\cite{SHD}) \\
    \textbf{SHD} & Spiking HD, obtained with Lauscher (from~\cite{SHD}) \\
    \textbf{S2S-HD} & spiked alternative to HD obtained with Speech2Spikes (from~\cite{speech2spikes})\\
    \textbf{NHD} & Neuromorphic HD, obtained with the HLP cochlear encoder (from~\cite{meunier_aicas2025})\\
    \hline
    \textbf{GSC} & Google Speech Commands v0.2 (from~\cite{Warden2018Speech}) \\
    \textbf{SSC} & Spiking GSC, obtained with Lauscher (from~\cite{SHD}) \\
    \textbf{S2S-GSC} & spiked alternative to GSC obtained with Speech2Spikes (from~\cite{speech2spikes})\\
    \textbf{NSC} & Neuromorphic GSC, obtained with the HLP cochlear encoder (from~\cite{meunier_aicas2025})\\
    \hline
    \textbf{NTIMIT} & Neuromorphic TIMIT, spiked alternative to TIMIT~\cite{TIMIT} obtained with our HLP cochlear encoder\\
    \hline
\end{tabular}

\section*{Acknowledgments}
This work is supported by a public grant overseen by the French National Research Agency (ANR) as part of the ‘Chaires IA’ programme (GrAI project ANR 19 CHIA 0003) and as part of the ‘PEPR IA France 2030’ programme (Emergences project ANR 23 PEIA 0002). This research is part of the programme DesCartes and is supported by the National Research Foundation, Prime Minister’s Office, Singapore under its Campus for Research Excellence and Technological Enterprise (CREATE) programme.


\bibliographystyle{IEEEtran}
\bibliography{references} 

@ARTICLE{SHD,
  author={Cramer, Benjamin and Stradmann, Yannik and Schemmel, Johannes and Zenke, Friedemann},
  journal={IEEE Transactions on Neural Networks and Learning Systems}, 
  title={The {Heidelberg} Spiking Data Sets for the Systematic Evaluation of Spiking Neural Networks}, 
  year={2022},
  volume={33},
  number={7},
  pages={2744-2757},
  doi={10.1109/TNNLS.2020.3044364}}

@misc{Warden2018Speech,
      title={Speech Commands: A Dataset for Limited-Vocabulary Speech Recognition}, 
      author={Pete Warden},
      year={2018},
      eprint={1804.03209},
      archivePrefix={arXiv},
      primaryClass={cs.CL},
      url={https://arxiv.org/abs/1804.03209}, 
}

@ARTICLE{N-MNIST,
AUTHOR={Orchard, Garrick  and Jayawant, Ajinkya  and Cohen, Gregory K.  and Thakor, Nitish },
TITLE={Converting Static Image Datasets to Spiking Neuromorphic Datasets Using Saccades},
JOURNAL={Frontiers in Neuroscience},
VOLUME={9},
YEAR={2015},
URL={https://www.frontiersin.org/journals/neuroscience/articles/10.3389/fnins.2015.00437},
DOI={10.3389/fnins.2015.00437},
ISSN={1662-453X}}

@article{Papierclemenceadrien,
  TITLE = {{A High-Level Methodology to Evaluate and Optimize Digital Architectures Targeting Spike Encoding}},
  AUTHOR = {Gillet, Cl{\'e}mence and Vincent, Adrien and Le Gal, Bertrand and Sa{\"i}ghi, Sylvain},
  URL = {https://hal.science/hal-04406432},
  JOURNAL = {{IEEE Access}},
  PUBLISHER = {{IEEE}},
  VOLUME = {11},
  PAGES = {120654-120665},
  YEAR = {2023},
  DOI = {10.1109/ACCESS.2023.3324877},
  HAL_ID = {hal-04406432},
  HAL_VERSION = {v1},
}

@article{audioOverview,
  title={Overview: Speech recognition technology, mel-frequency cepstral coefficients ({MFCC}), artificial neural network ({ANN})},
  author={Mistry, Divyesh S and Kulkarni, AV},
  journal={International Journal of Engineering Research and Technology},
  volume={2},
  number={10},
  year={2013},
  publisher={ESRSA Publications}
}

@article{hearingRange,
  title={Hearing ranges of laboratory animals},
  author={Heffner, Henry E and Heffner, Rickye S},
  journal={Journal of the American Association for Laboratory Animal Science},
  volume={46},
  number={1},
  pages={20--22},
  year={2007},
  publisher={American Association for Laboratory Animal Science}
}

@misc{MFCC,
      title={Choice of {Mel} Filter Bank in Computing {MFCC} of a Resampled Speech}, 
      author={Laxmi Narayana M. and Sunil Kumar Kopparapu},
      year={2014},
      eprint={1410.6903},
      archivePrefix={arXiv},
      primaryClass={cs.SD},
      url={https://arxiv.org/abs/1410.6903}, 
}

@ARTICLE{ENERGYDampfhoffer,
  author={Dampfhoffer, Manon and Mesquida, Thomas and Valentian, Alexandre and Anghel, Lorena},
  journal={IEEE Transactions on Emerging Topics in Computational Intelligence}, 
  title={Are {SNNs} Really More Energy-Efficient Than {ANNs}? an In-Depth Hardware-Aware Study}, year={2023},
  volume={7},
  number={3},
  pages={731-741},
  doi={10.1109/TETCI.2022.3214509}}

@INPROCEEDINGS{NbrOperationDampfhoffer,
  author={Dampfhoffer, Manon and Mesquida, Thomas and Hardy, Emmanuel and Valentian, Alexandre and Anghel, Lorena},
  booktitle={ICASSP 2023 - 2023 IEEE International Conference on Acoustics, Speech and Signal Processing (ICASSP)}, 
  title={Leveraging Sparsity with Spiking Recurrent Neural Networks for Energy-Efficient Keyword Spotting}, 
  year={2023},
  volume={},
  number={},
  pages={1-5},
  doi={10.1109/ICASSP49357.2023.10097174}}

@INPROCEEDINGS{LIF,
  author={Rast, A. D. and Galluppi, F. and Jin, X. and Furber, S.B.},
  booktitle={The 2010 International Joint Conference on Neural Networks (IJCNN)}, 
  title={The Leaky Integrate-and-Fire neuron: A platform for synaptic model exploration on the {SpiNNaker} chip}, 
  year={2010},
  volume={},
  number={},
  pages={1-8},
  doi={10.1109/IJCNN.2010.5596364}}

@article{Chan2007AEREAR,
  title={{AER EAR}: A matched silicon cochlea pair with address event representation interface},
  author={Chan, Vincent and Liu, Shih-Chii and van Schaik, Andr},
  journal={IEEE Transactions on Circuits and Systems I: Regular Papers},
  volume={54},
  number={1},
  pages={48--59},
  year={2007},
  publisher={IEEE}
}

@inproceedings{v2e,
  title={{v2e}: From video frames to realistic {DVS} events},
  author={Hu, Yuhuang and Liu, Shih-Chii and Delbruck, Tobi},
  booktitle={Proceedings of the IEEE/CVF conference on computer vision and pattern recognition},
  pages={1312--1321},
  year={2021}
}

@INPROCEEDINGS{DVSGesture,
  author={Amir, Arnon and Taba, Brian and Berg, David and Melano, Timothy and McKinstry, Jeffrey and Di Nolfo, Carmelo and Nayak, Tapan and Andreopoulos, Alexander and Garreau, Guillaume and Mendoza, Marcela and Kusnitz, Jeff and Debole, Michael and Esser, Steve and Delbruck, Tobi and Flickner, Myron and Modha, Dharmendra},
  booktitle={2017 IEEE Conference on Computer Vision and Pattern Recognition ({CVPR})}, 
  title={A Low Power, Fully Event-Based Gesture Recognition System}, 
  year={2017},
  volume={},
  number={},
  pages={7388-7397},
  doi={10.1109/CVPR.2017.781}
}

@article{Gen1,
  author       = {Pierre de Tournemire and
                  Davide Nitti and
                  Etienne Perot and
                  Davide Migliore and
                  Amos Sironi},
  title        = {A Large Scale Event-based Detection Dataset for Automotive},
  journal      = {CoRR},
  volume       = {abs/2001.08499},
  year         = {2020},
  url          = {https://arxiv.org/abs/2001.08499},
  eprinttype    = {arXiv},
  eprint       = {2001.08499},
  bibsource    = {dblp computer science bibliography, https://dblp.org}
}

@misc{v2ce,
      title={{V2CE}: Video to Continuous Events Simulator}, 
      author={Zhongyang Zhang and Shuyang Cui and Kaidong Chai and Haowen Yu and Subhasis Dasgupta and Upal Mahbub and Tauhidur Rahman},
      year={2024},
      eprint={2309.08891},
      archivePrefix={arXiv},
      primaryClass={cs.CV},
      url={https://arxiv.org/abs/2309.08891}, 
}

@InProceedings{vid2e,
  author = {Daniel Gehrig and Mathias Gehrig and Javier Hidalgo-Carri\'o and Davide Scaramuzza},
  title = {Video to Events: Recycling Video Datasets for Event Cameras},
  booktitle = {{IEEE} Conf. Comput. Vis. Pattern Recog. (CVPR)},
  month = {June},
  year = {2020}
}

@article{Yang2023,
title = {Neuromorphic electronics for robotic perception, navigation and control: A survey},
journal = {Engineering Applications of Artificial Intelligence},
volume = {126},
pages = {106838},
year = {2023},
issn = {0952-1976},
doi = {https://doi.org/10.1016/j.engappai.2023.106838},
url = {https://www.sciencedirect.com/science/article/pii/S0952197623010229},
author = {Yi Yang and Chiara Bartolozzi and Haiyan H. Zhang and Robert A. Nawrocki}
}

@article{Vanarse2016,
  title={A review of current neuromorphic approaches for vision, auditory, and olfactory sensors},
  author={Vanarse, Anup and Osseiran, Adam and Rassau, Alexander},
  journal={Frontiers in Neuroscience},
  volume={10},
  pages={115},
  year={2016},
  publisher={Frontiers Media SA}
}

@inproceedings{speech2spikes,
    author = {Stewart, Kenneth Michael and Shea, Timothy and Pacik-Nelson, Noah and Gallo, Eric and Danielescu, Andreea},
    title = {{Speech2Spikes}: Efficient Audio Encoding Pipeline for Real-time Neuromorphic Systems},
    year = {2023},
    isbn = {9781450399470},
    publisher = {Association for Computing Machinery},
    address = {New York, NY, USA},
    url = {https://doi.org/10.1145/3584954.3584995},
    doi = {10.1145/3584954.3584995},
    booktitle = {Proceedings of the 2023 Annual Neuro-Inspired Computational Elements Conference},
    pages = {71–78},
    numpages = {8},
    location = {San Antonio, TX, USA},
    series = {NICE '23}
}

@article{TIMIT,
  title={TIMIT acoustic-phonetic continuous speech corpus},
  author={Garofolo, John S and Lamel, Lori F and Fisher, William M and Pallett, David S and Dahlgren, Nancy L and Zue, Victor and Fiscus, Jonathan G},
  journal={(No Title)},
  year={1993},
  publisher={Linguistic data consortium}
}

@article{ElFerdaoussi2022,
  author    = {El Ferdaoussi, Ahmad and Plourde, Eric and Rouat, Jean},
  title     = {Evaluation of Neuromorphic Spike Encoding of Sound Using Information Theory},
  journal   = {arXiv preprint arXiv:2202.09619},
  year      = {2022},
  doi       = {10.48550/arXiv.2202.09619}
}

@article{HTSpike2025,
  author    = {Haghighatshoar, Saeid and Muir, Dylan R.},
  title     = {Low-power Spiking Neural Network Audio Source Localisation Using a Hilbert Transform Audio Event Encoding Scheme},
  journal   = {Communications Engineering},
  volume    = {4},
  pages     = {18},
  year      = {2025},
  doi       = {10.1038/s44172-025-00359-9}
}

@article{EffSpikeSpeech2022,
  author    = {Yarga, Sidi Yaya Arnaud and Rouat, Jean and Wood, Sean U. N.},
  title     = {Efficient Spike Encoding Algorithms for Neuromorphic Speech Recognition},
  journal   = {arXiv preprint arXiv:2207.07073},
  year      = {2022}
}

@article{EncodingSurvey2021,
  author    = {Auge, Daniel and Hille, Julian and Müller, Etienne},
  title     = {A Survey of Encoding Techniques for Signal Processing in Spiking Neural Networks},
  journal   = {Neural Processing Letters},
  volume    = {54},
  number    = {4},
  pages     = {567--589},
  year      = {2021},
  doi       = {10.1007/s11063-021-10562-2}
}

@article{SurveyAudio2025,
  author    = {Basu, Amlan and Chaudhari, Pranav and Di Caterina, Gaetano},
  title     = {Fundamental Survey on Neuromorphic Based Audio Classification},
  journal   = {arXiv preprint arXiv:2502.15056},
  year      = {2025}
}

@article{lee2002speaker,
  title={Speaker-independent phone recognition using hidden Markov models},
  author={Lee, K-F and Hon, H-W},
  journal={IEEE Transactions on acoustics, speech, and signal processing},
  volume={37},
  number={11},
  pages={1641--1648},
  year={2002},
  publisher={IEEE}
}

@inproceedings{schone2024scalable,
  title={Scalable event-by-event processing of neuromorphic sensory signals with deep state-space models},
  author={Sch{\"o}ne, Mark and Sushma, Neeraj Mohan and Zhuge, Jingyue and Mayr, Christian and Subramoney, Anand and Kappel, David},
  booktitle={2024 International Conference on Neuromorphic Systems (ICONS)},
  pages={124--131},
  year={2024},
  organization={IEEE}
}

@article{baronig2024advancing,
  title={Advancing spatio-temporal processing in spiking neural networks through adaptation},
  author={Baronig, Maximilian and Ferrand, Romain and Sabathiel, Silvester and Legenstein, Robert},
  journal={arXiv preprint arXiv:2408.07517},
  year={2024}
}

@inproceedings{patino2023empirical,
  title={Empirical study on the efficiency of spiking neural networks with axonal delays, and algorithm-hardware benchmarking},
  author={Pati{\~n}o-Saucedo, Alberto and Yousefzadeh, Amirreza and Tang, Guangzhi and Corradi, Federico and Linares-Barranco, Bernab{\'e} and Sifalakis, Manolis},
  booktitle={2023 IEEE International Symposium on Circuits and Systems (ISCAS)},
  pages={1--5},
  year={2023},
  organization={IEEE}
}

@article{bittar2022surrogate,
  title={A surrogate gradient spiking baseline for speech command recognition},
  author={Bittar, Alexandre and Garner, Philip N},
  journal={Frontiers in Neuroscience},
  volume={16},
  pages={865897},
  year={2022},
  publisher={Frontiers Media SA}
}

@article{sun2025pfa,
  title   = {Towards parameter-free attentional spiking neural networks},
  author  = {Sun, Pengfei and Wu, Jibin and Devos, Paul and Botteldooren, Dick},
  journal = {Neural Networks},
  year    = {2025},
  volume  = {185},
  pages   = {107154},
  doi     = {10.1016/j.neunet.2025.107154},
  url     = {https://doi.org/10.1016/j.neunet.2025.107154}
}

@article{baronig2025adlifsnn,
  title   = {Advancing spatio-temporal processing through adaptation in spiking neural networks},
  author  = {Baronig, Maximilian and Ferrand, Romain and Sabathiel, Silvester and Legenstein, Robert},
  journal = {Nature Communications},
  year    = {2025},
  volume  = {16},
  number  = {1},
  pages   = {5776},
  doi     = {10.1038/s41467-025-60878-z},
  url     = {https://doi.org/10.1038/s41467-025-60878-z}
}

@article{xu2025asrcsnn,
  title         = {ASRC-SNN: Adaptive Skip Recurrent Connection Spiking Neural Network},
  author        = {Xu, Shang and Zhang, Jiayu and Wang, Ziming and Jiang, Runhao and Yan, Rui and Tang, Huajin},
  journal       = {arXiv preprint arXiv:2505.11455},
  year          = {2025},
  url           = {https://arxiv.org/abs/2505.11455},
  eprint        = {2505.11455},
  archivePrefix = {arXiv},
  primaryClass  = {cs.NE}
}

@article{deckers2024co,
  title={Co-learning synaptic delays, weights and adaptation in spiking neural networks},
  author={Deckers, Lucas and Van Damme, Laurens and Van Leekwijck, Werner and Tsang, Ing Jyh and Latr{\'e}, Steven},
  journal={Frontiers in Neuroscience},
  volume={18},
  pages={1360300},
  year={2024},
  publisher={Frontiers Media SA}
}

@inproceedings{chakrabartty2010exploiting,
  title={Exploiting spike-based dynamics in a silicon cochlea for speaker identification},
  author={Chakrabartty, Shantanu and Liu, Shih-Chii},
  booktitle={Proceedings of 2010 IEEE International Symposium on Circuits and Systems},
  pages={513--516},
  year={2010},
  organization={IEEE}
}

@inproceedings{li2012real,
  title={Real-time speaker identification using the AEREAR2 event-based silicon cochlea},
  author={Li, Cheng-Han and Delbruck, Tobi and Liu, Shih-Chii},
  booktitle={2012 IEEE international symposium on circuits and systems (ISCAS)},
  pages={1159--1162},
  year={2012},
  organization={IEEE}
}

@article{pan2020efficient,
  title={An efficient and perceptually motivated auditory neural encoding and decoding algorithm for spiking neural networks},
  author={Pan, Zihan and Chua, Yansong and Wu, Jibin and Zhang, Malu and Li, Haizhou and Ambikairajah, Eliathamby},
  journal={Frontiers in neuroscience},
  volume={13},
  pages={1420},
  year={2020},
  publisher={Frontiers Media SA}
}

@article{dong2018unsupervised,
  title={Unsupervised speech recognition through spike-timing-dependent plasticity in a convolutional spiking neural network},
  author={Dong, Meng and Huang, Xuhui and Xu, Bo},
  journal={PloS one},
  volume={13},
  number={11},
  pages={e0204596},
  year={2018},
  publisher={Public Library of Science San Francisco, CA USA}
}

@article{maass1997networks,
  title={Networks of spiking neurons: the third generation of neural network models},
  author={Maass, Wolfgang},
  journal={Neural networks},
  volume={10},
  number={9},
  pages={1659--1671},
  year={1997},
  publisher={Elsevier}
}

@book{gerstner2002snm,
  title     = {Spiking Neuron Models: Single Neurons, Populations, Plasticity},
  author    = {Gerstner, Wulfram and Kistler, Werner M.},
  year      = {2002},
  publisher = {Cambridge University Press}
}

@article{bal2025pspikessm,
  title={P-spikessm: Harnessing probabilistic spiking state space models for long-range dependency tasks},
  author={Bal, Malyaban and Sengupta, Abhronil},
  journal={arXiv preprint arXiv:2406.02923},
  year={2024}
}

@article{eshraghian2021training,
  title   = {Training spiking neural networks using lessons from deep learning},
  author  = {Eshraghian, Jason K and Ward, Max and Neftci, Emre and Wang, Xinxin
             and Lenz, Gregor and Dwivedi, Girish and Bennamoun, Mohammed and
             Jeong, Doo Seok and Lu, Wei D},
  journal = {Proceedings of the IEEE},
  volume  = {111},
  number  = {9},
  pages   = {1016--1054},
  year    = {2023}
}

@article{cramer2019heidelberg,
  title={The Heidelberg spiking datasets for the systematic evaluation of spiking neural networks},
  author={Cramer, Benjamin and Stradmann, Yannik and Schemmel, Johannes and Zenke, Friedemann},
  journal={arXiv preprint arXiv:1910.07407},
  year={2019}
}

@article{wang2024spikescr,
  title={Efficient Speech Command Recognition Leveraging Spiking Neural Network and Curriculum Learning-based Knowledge Distillation},
  author={Wang, Jiaqi and Yu, Liutao and Huang, Liwei and Zhou, Chenlin and Zhang, Han and Song, Zhenxi and Zhang, Min and Ma, Zhengyu and Zhang, Zhiguo},
  journal={arXiv preprint arXiv:2412.12858},
  year={2024}
}

@inproceedings{hammouamri2024dcls,
  title={Learning Delays in Spiking Neural Networks using Dilated Convolutions with Learnable Spacings},
  author={Hammouamri, Ismail and Khalfaoui-Hassani, Idder and Masquelier, Timoth{\'e}e},
  booktitle={International Conference on Learning Representations (ICLR)},
  year={2024},
  note={OpenReview: \url{https://openreview.net/forum?id=4r2ybzJnmN}}
}

@inproceedings{Song2024Spiking,
  title={{Spiking-Leaf}: A Learnable Auditory Front-End for Spiking Neural Networks},
  author={Song, Zeyang and Wu, Jibin and Zhang, Malu and Shou, Mike Zheng and Li, Haizhou},
  booktitle={ICASSP 2024-2024 IEEE International Conference on Acoustics, Speech and Signal Processing (ICASSP)},
  pages={226--230},
  year={2024},
  organization={IEEE}
}

@article{glasberg1990derivation,
  title={Derivation of auditory filter shapes from notched-noise data},
  author={Glasberg, Brian R. and Moore, Brian C. J.},
  journal={The Journal of the Acoustical Society of America},
  volume={87},
  number={2},
  pages={861--866},
  year={1990}
}

@techreport{slaney1998auditory,
  title={Auditory Toolbox, Version 2},
  author={Slaney, Malcolm},
  institution={Interval Research Corporation},
  number={1998-010},
  year={1998},
  note={Includes mel and gammatone filter-bank implementations}
}

@INPROCEEDINGS{meunier_aicas2025,
  author={Meunier, Valentin and Gruel, Amélie and Vincent, Adrien F. and Saïghi, Sylvain},
  booktitle={2025 IEEE 7th International Conference on Artificial Intelligence Circuits and Systems (AICAS)}, 
  title={Comparison of Hardware-friendly, Audio-to-spikes Cochlear Encoding for Neuromorphic Processing}, 
  year={2025},
  volume={},
  number={},
  pages={1-5},
  doi={10.1109/AICAS64808.2025.11173106}}

@article{sorbaro2019,
  title={Optimizing the Energy Consumption of Spiking Neural Networks for Neuromorphic Applications},
  author={M. Sorbaro and Li-Yu Daisy Liu and Massimo Bortone and Sadique Sheik},
  journal={Front. in Neuroscience},
  year={2019},
  volume={14},
  url={https://api.semanticscholar.org/CorpusID:208547817}
}

@incollection{filter_order,
    title = {2 - Biomedical signal processing technique},
    editor = {Suman Lata Tripathi and Valentina Emilia Balas and Mufti Mahmud and Soumya Banerjee},
    booktitle = {Machine Learning Models and Architectures for Biomedical Signal Processing},
    publisher = {Academic Press},
    pages = {19-42},
    year = {2025},
    isbn = {978-0-443-22158-3},
    doi = {https://doi.org/10.1016/B978-0-443-22158-3.00002-8},
    url = {https://www.sciencedirect.com/science/article/pii/B9780443221583000028},
    author = {Manoj Singh Adhikari and Manoj Sindhwani and Shippu Sachdeva}
}

@InCollection{Shrestha2018,
  author    = {Shrestha, Sumit Bam and Orchard, Garrick},
  title     = {{SLAYER}: Spike Layer Error Reassignment in Time},
  booktitle = {Advances in Neural Information Processing Systems 31},
  publisher = {Curran Associates, Inc.},
  year      = {2018},
  editor    = {S. Bengio and H. Wallach and H. Larochelle and K. Grauman and N. Cesa-Bianchi and R. Garnett},
  pages     = {1419--1428},
  url       = {http://papers.nips.cc/paper/7415-slayer-spike-layer-error-reassignment-in-time.pdf},
}

@misc{adamoptimiser,
    title={Adam: A Method for Stochastic Optimization}, 
    author={Diederik P. Kingma and Jimmy Ba},
    year={2017},
    eprint={1412.6980},
    archivePrefix={arXiv},
    primaryClass={cs.LG},
    url={https://arxiv.org/abs/1412.6980}, 
}

@article{Mead1988,
  title={A silicon model of early visual processing},
  author={Mead, Carver A and Mahowald, Misha A},
  journal={Neural networks},
  volume={1},
  number={1},
  pages={91--97},
  year={1988},
  publisher={Elsevier}
}

@article{Mead2020,
  title={How we created neuromorphic engineering},
  author={Mead, Carver},
  journal={Nature Electronics},
  volume={3},
  number={7},
  pages={434--435},
  year={2020},
  publisher={Nature Publishing Group UK London}
}

@article{Nowotny2025,
  title={Loss shaping enhances exact gradient learning with Eventprop in spiking neural networks},
  author={Nowotny, Thomas and Turner, James P and Knight, James C},
  journal={Neuromorphic Computing and Engineering},
  volume={5},
  number={1},
  pages={014001},
  year={2025}
}

@inproceedings{sun2023adaptive,
  title={Adaptive Axonal Delays in feedforward spiking neural networks for accurate spoken word recognition},
  author={Sun, Pengfei and Eqlimi, Ehsan and Chua, Yansong and Devos, Paul and Botteldooren, Dick},
  booktitle={ICASSP 2023-2023 IEEE International Conference on Acoustics, Speech and Signal Processing (ICASSP)},
  pages={1--5},
  year={2023},
  organization={IEEE}
}

@misc{queant2025delreclearningdelaysrecurrent,
      title={DelRec: learning delays in recurrent spiking neural networks}, 
      author={Alexandre Queant and Ulysse Rançon and Benoit R Cottereau and Timothée Masquelier},
      year={2025},
      eprint={2509.24852},
      archivePrefix={arXiv},
      primaryClass={cs.NE},
      url={https://arxiv.org/abs/2509.24852}, 
}

@misc{fabre2026,
      title={SiLIF: Structured State Space Model Dynamics and Parametrization for Spiking Neural Networks}, 
      author={Maxime Fabre and Lyubov Dudchenko and Younes Bouhadjar and Emre Neftci},
      year={2026},
      eprint={2506.06374},
      archivePrefix={arXiv},
      primaryClass={cs.NE},
      url={https://arxiv.org/abs/2506.06374}, 
}

@misc{xu2025,
      title={ASRC-SNN: Adaptive Skip Recurrent Connection Spiking Neural Network}, 
      author={Shang Xu and Jiayu Zhang and Ziming Wang and Runhao Jiang and Rui Yan and Huajin Tang},
      year={2025},
      eprint={2505.11455},
      archivePrefix={arXiv},
      primaryClass={cs.NE},
      url={https://arxiv.org/abs/2505.11455}, 
}

@INPROCEEDINGS{richter2025,
  author={Richter, Simon and Khatiboun, Darío Fernández and Sadeghi, Maryam and Zamani, Milad and Moradi, Farshad},
  booktitle={2025 IEEE 7th International Conference on Artificial Intelligence Circuits and Systems (AICAS)}, 
  title={Non-uniform Memory Partitioning For Low-Power Spiking Neural Networks}, 
  year={2025},
  volume={},
  number={},
  pages={1-5}
 }


\end{document}